\documentclass[10pt,twocolumn,letterpaper]{article}

\usepackage[pagenumbers]{cvpr} % To force page numbers, e.g. for an arXiv version

\usepackage{graphicx}
\usepackage{amsmath}
\usepackage{amssymb}
\usepackage{booktabs}

\usepackage{cases} %dc

\usepackage[dvipsnames]{xcolor} % dc
\usepackage[pagebackref,breaklinks,colorlinks]{hyperref}

\usepackage[capitalize]{cleveref}
\crefname{section}{Sec.}{Secs.}
\Crefname{section}{Section}{Sections}
\Crefname{table}{Table}{Tables}
\crefname{table}{Tab.}{Tabs.}

\def\cvprPaperID{401} % *** Enter the CVPR Paper ID here
\def\confName{CVPR}
\def\confYear{2022}

\newcommand{\eR}{\mathbb{R}}

\newcommand{\eAT}{A^{\dag}}

\newcommand{\eeAT}{A^{\dag}}

\newcommand{\nullA}{\mathcal{N}_A}

\newcommand{\ntransf}{|\mathcal{G}|}

\newtheorem{theorem}{\textbf{\emph{Theorem}}}

\newtheorem{lemma}{\textbf{\emph{Lemma}}}

\begin{document}

%%%%%%%%% TITLE - PLEASE UPDATE
\title{Robust Equivariant Imaging: a fully unsupervised framework for learning to image from noisy and partial measurements}
%Self-Supervised Learning From Noisy Measurement Data}

\author{Dongdong Chen\thanks{The first two authors have contributed equally to this paper.}\\
School of Engineering\\
University of Edinburgh\\
{\tt\small d.chen@ed.ac.uk}\and
Julián Tachella$^\ast$\\
School of Engineering\\
University of Edinburgh\\
{\tt\small julian.tachella@ed.ac.uk}
\and
Mike E. Davies\\
School of Engineering\\
University of Edinburgh\\
{\tt\small mike.davies@ed.ac.uk}
}
\maketitle

%%%%%%%%% ABSTRACT
\begin{abstract}
Deep networks provide state-of-the-art performance in multiple imaging inverse problems ranging from medical imaging to computational photography. However, most existing networks are trained with clean signals which are often hard or impossible to obtain. Equivariant imaging (EI) is a recent self-supervised learning framework that exploits the group invariance  present in signal distributions to learn a reconstruction function from partial measurement data alone. While EI results are impressive, its performance degrades with increasing noise. In this paper, we propose a Robust Equivariant Imaging (REI) framework which  can learn to image from noisy partial measurements alone. The proposed method uses Stein's Unbiased Risk Estimator (SURE) to obtain a fully  unsupervised training loss that is robust to noise. We show that REI leads to considerable performance gains on linear and nonlinear inverse problems, thereby paving the way for robust unsupervised imaging with deep networks. Code is available at \url{https://github.com/edongdongchen/REI}.
\end{abstract}

%%%%%%%%% BODY TEXT
\section{Introduction}
\label{sec:intro}
Inverse problems play a fundamental role in  computer vision and signal processing. Applications such as computed tomography (CT), magnetic resonance imaging (MRI), super-resolution and image inpainting have been extensively explored. The goal in an inverse problem is to reconstruct a signal $x\in\eR^n$ from measurements $y\in\eR^m$, that is inverting the forward process
\begin{equation}
    y = A(x) + \epsilon,
\end{equation}
which is generally a challenging task due to the noise $\epsilon$ and the ill-conditioned or rank deficient forward operator $A$ (e.g., $m< n$). Traditional approaches exploit prior knowledge about $x$ (sparsity~\cite{gleichman2011blind}, total variation~\cite{rudin1992nonlinear} or more recently the deep image prior~\cite{ulyanov2018deep}) to regularize the reconstructions. A  different approach is taken by more recent end-to-end learning solutions~\cite{jin2017deep} which aim to learn the inverse mapping directly from $(x,y)$ pairs. However, in many inverse problems, it is expensive or impossible to obtain clean signals $x$, and only noisy measurement data $y$ are available for training.

\begin{figure}[t]
\begin{minipage}{1\linewidth}
\centerline{
\includegraphics[width=1\textwidth]{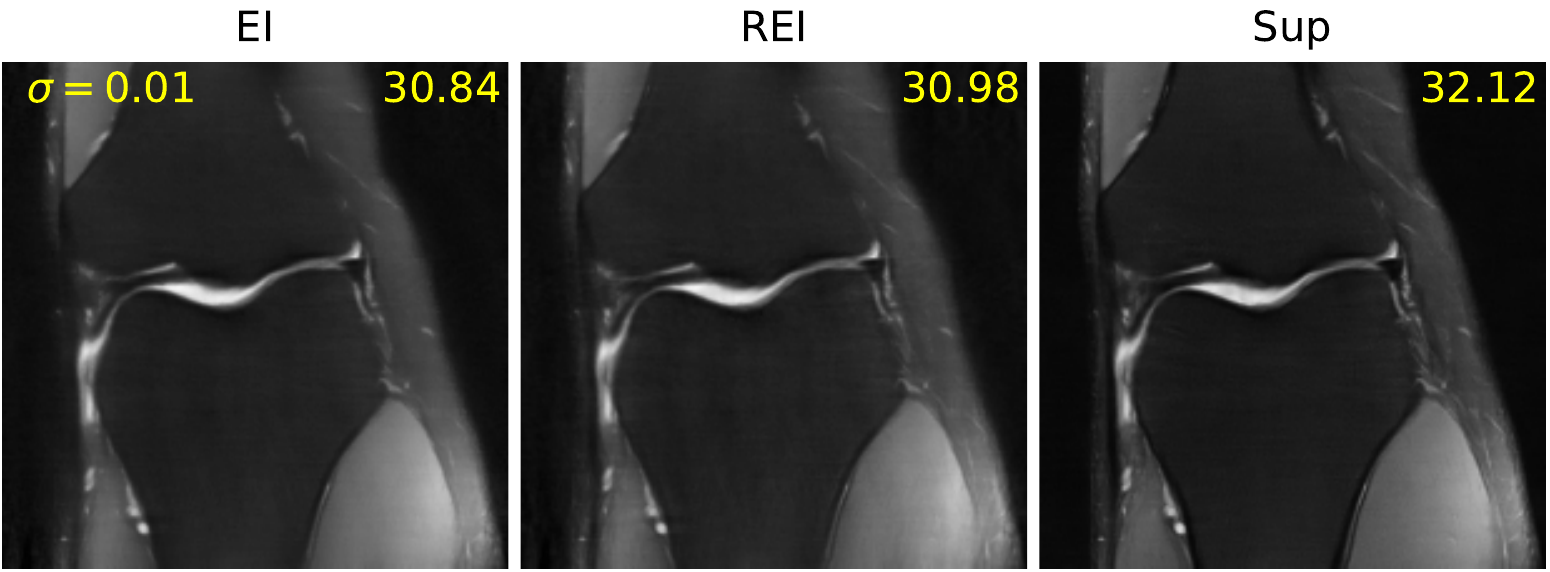}}
\end{minipage}
\begin{minipage}{1\linewidth}
\centerline{
\includegraphics[width=1\textwidth]{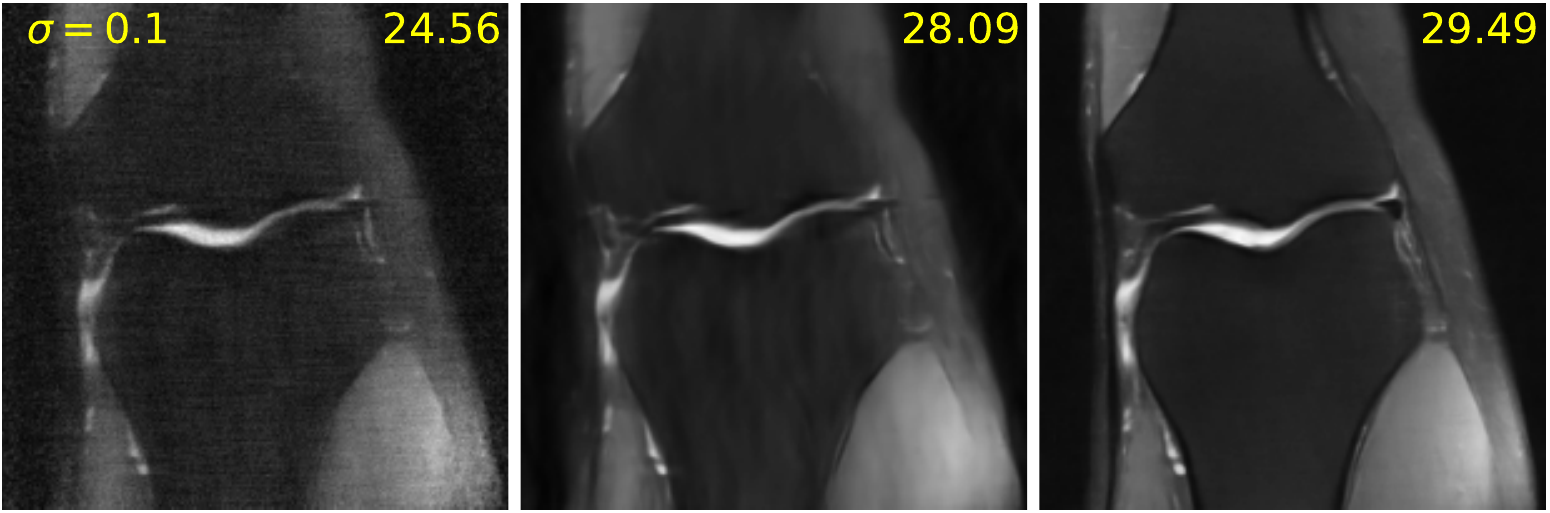}}
\end{minipage}
\begin{minipage}{1\linewidth}
\centerline{
\includegraphics[width=1\textwidth]{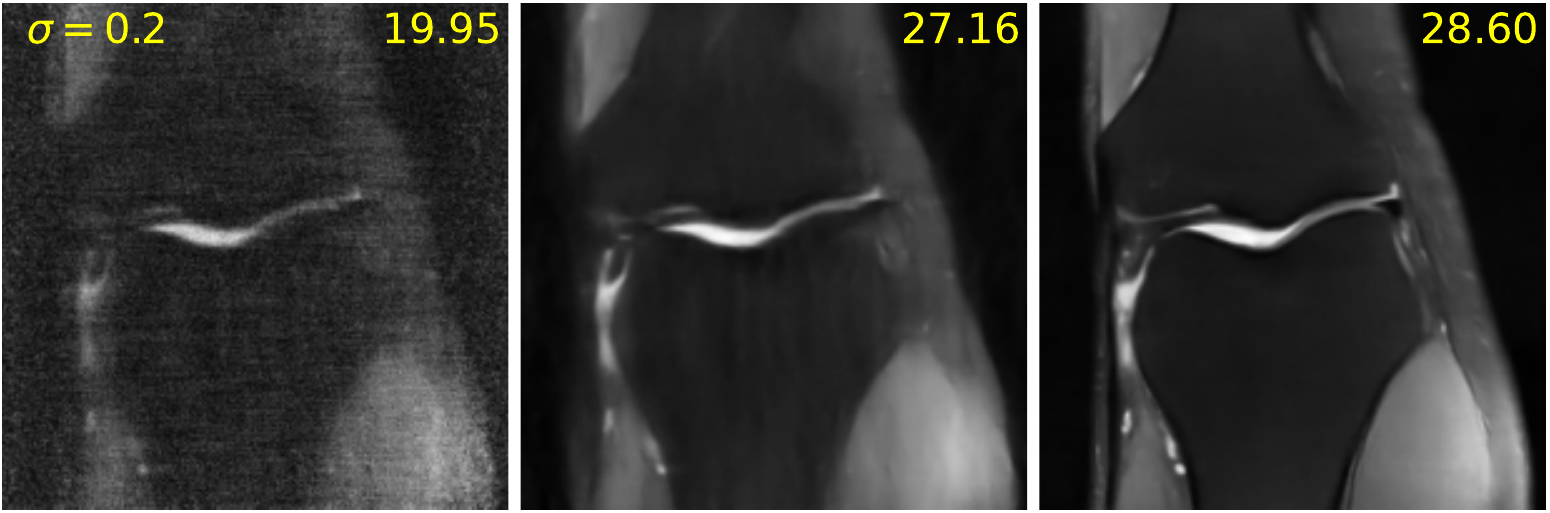}}
\end{minipage}
\caption{\textbf{REI vs EI}: EI is not robust to noise and the performance degrades with increasing noise. From top to bottom: reconstructions of EI, supervised (Sup) baseline, and the proposed REI on $4\times$ accelerated MRI with Gaussian noise level $\sigma=0.01, 0.1, 0.2$. PSNR values are shown in the top right corner of the images.}
\label{fig:motivation}
\end{figure}
% \vspace{-5pt}

Without any assumptions on the signal distribution, it is fundamentally impossible to learn the reconstruction mapping from partial measurements alone, as there are infinitely many inverses for each observed measurement $y$~\cite{gleichman2011blind,chen2021equivariant}. Some unsupervised methods alleviate this limitation by using multiple forward operators~\cite{bora2018ambientgan,pajot2019unsupervised,gan2021deep}, however this setting is not often met in practice.
Recently, the EI framework~\cite{chen2021equivariant} showed that learning from partial measurements $y$ is possible if the underlying signal set is invariant to transformations, such as shifts or rotations. The EI strategy offers an elegant way to learn the model in a self-supervised fashion, as most natural signals present certain invariances (e.g., invariance to translations and rotations in images).
EI was shown to achieve similar performance to supervised methods on a number of noiseless image reconstruction tasks such as sparse-view CT and image inpainting. However, as shown in~\Cref{fig:motivation}, EI's performance degrades rapidly in the presence of measurement noise.

In this paper, we propose a novel robust EI (REI) framework which can learn from only noisy measurements data. In particular, we use Stein's Unbiased Risk Estimator (SURE)~\cite{stein1981estimation} to derive an unsupervised loss that takes into account the noise present in the measurements. The proposed learning method is architecture agnostic and can handle a wide range of different noise models, including Gaussian noise, Poisson noise and mixed Poisson-Gaussian noise. We experimentally show that our REI method outperforms the vanilla EI for all noise levels in 3 inverse problems: sparse-view CT, accelerated MRI and image inpainting. In summary, our contributions are as follows:

\begin{enumerate}
\item We present a new end-to-end unsupervised learning framework for training deep networks to solve imaging problems from noisy partial measurements alone.
\item Our REI learning framework can handle a wide range of the most frequently encountered noise models in practice, including: Gaussian noise, Poisson noise, and mixed Gaussian-Poisson noise.

\item We evaluate the proposed method on accelerated MRI,  image inpainting, and a nonlinear imaging model for sparse-view CT, showing substantial improvements over vanilla EI in the presence of noise and approaching the quality of reconstructions trained with ground truth data using supervised learning.
\end{enumerate}

\section{Background}
We begin with some basic definitions. Let  $A:\eR^{n} \rightarrow \eR^{m}$ be a (possibly nonlinear) smooth measurement operator with constant rank $m<n$. For a given measurement vector $y\in\eR^{m}$, the set of feasible solutions $A^{-1}(y)=\{ x\in \eR^{n} | A(x)=y \}$ is a submanifold of $\eR^{n}$ of dimension $n-m$~\cite[Chapter~4]{lee2013smooth}.

\subsection{Learning to Solve Inverse Problems} \label{learning to solve inv prob}
Modern learning-based approaches learn a reconstruction function $f_\theta:\mathbb{R}^{m}\rightarrow\mathbb{R}^{n}$ which maps the observed measurements $y$ to the reconstructed signal $x$. The function is generally parameterized by a convolutional network and trained using $N$ pairs of measurements and associated reconstructions $\{(y_i,x_i)\}_{i=1,\dots,N}$. Learning only with partial measurements $y_i$ is impossible without any further assumptions~\cite{chen2021equivariant,gleichman2011blind} as $f_\theta(y)$ can output any of the $x\in A^{-1}(y)$ and still perfectly fit the training set.

\begin{figure}[t]
\includegraphics[width=1\linewidth]{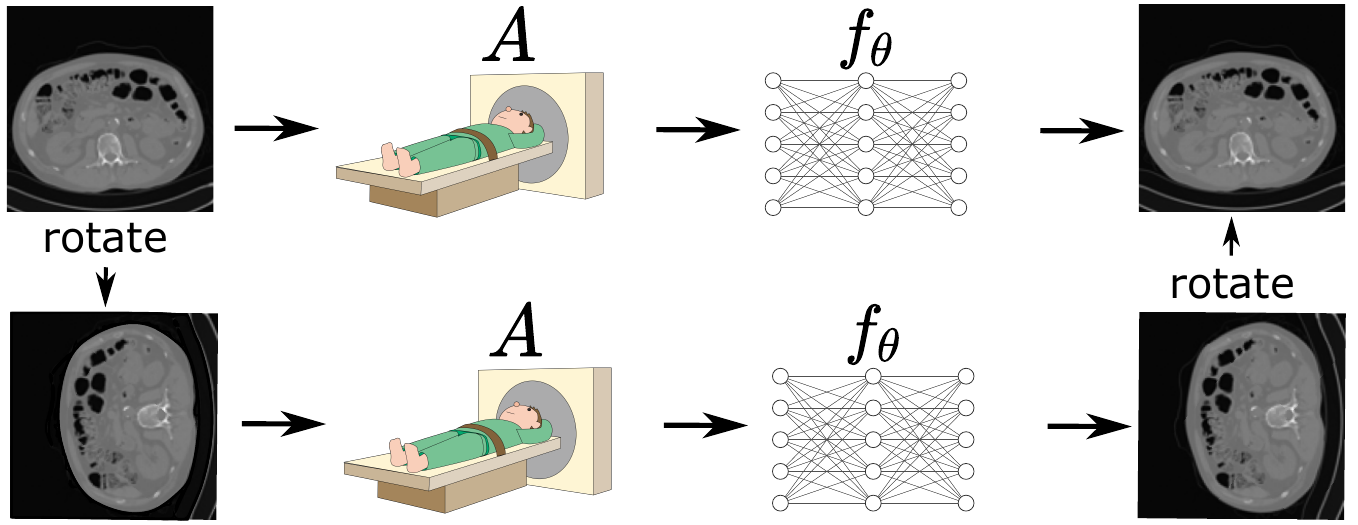}
\caption{\textbf{Equivariant imaging systems.} If the set of signals is invariant to a certain set of transformations, the composition of imaging operator $A$ with the reconstruction function $f_\theta$ should be equivariant to these transformations.}
\label{fig:equiv system}
\end{figure}

\paragraph{Equivariant Imaging (EI)}
Recently, the EI framework~\cite{chen2021equivariant} showed that learning  with only measurement data $y_i$ is possible if the signal set is assumed to be group invariant. That is, the signal set $\mathcal{X}\subseteq\mathbb{R}^{n}$ is invariant to a group of transformations (e.g., shifts, rotations, etc.) $\{T_g\}_{g=1,\dots,\ntransf}$ if for all $x\in\mathcal{X}$, $T_gx$ also belongs to $\mathcal{X}$ for all transformations $g=1,\dots,\ntransf$. Assuming that the transformation matrices are unitary\footnote{This can always be obtained after a change of basis if the group is compact, see for example~\cite{kondor2008group}.} and there is no noise affecting the measurements,
for all group elements $g=1,\dots,\ntransf$, we have that \begin{equation}
    y=A(x)=A(T_gT_g^{\top}x)=A_g(\tilde{x})
\end{equation}
where $A_g = A\circ T_g$ and both $x$ and $\tilde{x}=T_g^{\top}x$ belong to the signal set $\mathcal{X}$ due to the invariance property. Thus, the invariance property gives access to multiple \emph{virtual} operators $\{A_g\}_{g=1,\dots,\ntransf}$. For a given measurement $y$, each operator $A_g$ might have a different submanifold of possible inverses $A_g^{-1}(y)$. Thus, it is feasible to learn $f_\theta$ from only measurements if the intersection of these submanifolds $\bigcap_g A_g^{-1}(y)$ contains a single signal $x$.

The invariance property also tells us that the composition $f_{\theta}\circ A$ should be equivariant to the transformations (see Figure~\ref{fig:equiv system}): %(see~\cref{fig:equiv system}):
\begin{equation}\label{eq:equivariant cond}
    T_gf_{\theta}\left(A(x)\right) = f_{\theta}\left(A(T_gx) \right)
\end{equation}
for $g=1,\dots,\ntransf$. To exploit this idea the EI training strategy enforces both measurement consistency $A(f_{\theta}(y))=y$ and the equivariance condition in~\eqref{eq:equivariant cond} using a dataset of measurements $\{y_i\}_{i=1,\dots,N}$. The reconstruction function $f_{\theta}$ is then learned by minimizing the following loss:
\begin{equation}\label{eqs:ten_loss}
      \mathcal{L}_{\text{EI}} (\theta) =   \mathcal{L}_{\text{MC}}(\theta) + \alpha \mathcal{L}_{\text{EQ}} (\theta) ,
\end{equation}
with
\begin{align}
\mathcal{L}_{\text{MC}} (\theta)&= \sum_{i=1}^{N}\frac{1}{m}\|y_i-A\left(f_\theta(y_i)\right)\|^2\label{eqs:loss_mc},\\
\mathcal{L}_{\text{EQ}}(\theta)&=\sum_{i=1}^{N}\sum_{g=1}^{\ntransf}\frac{1}{\ntransf n}  \|T_gf_\theta(y_i)-f_\theta\left(A(T_gf_\theta(y_i))\right)\|^2\label{eqs:loss_eq}
\end{align}
where the first term \eqref{eqs:loss_mc} enforces measurement consistency (MC) and the second term \eqref{eqs:loss_eq} enforces system equivariance, and $\alpha$ is a trade-off parameter.

While EI proved to be an efficient self-supervised learning method in noiseless settings~\cite{chen2021equivariant}, its performance degrades rapidly in the presence of noise, as shown in Figure~\ref{fig:motivation}, hindering its use in noisy inverse problems.

\subsection{Self-supervised denoising via SURE} \label{subsec: sure}

In most real-world settings, observed measurements are corrupted by noise. For example, measurements can contain Gaussian noise, that is
\begin{equation}
    y|u \sim \mathcal{N} (u, I\sigma^2),
\end{equation}
where $u$ denotes the clean measurements, $\sigma$ is the standard deviation of the noise and $I$ denotes the identity matrix.
A denoiser network $h_\theta$ that maps $y\mapsto u$ is generally trained with a dataset of noisy and clean pairs $(y,u)$ via the supervised loss
\begin{equation}
  \mathcal{L}_{\text{MSE}} (\theta) = \frac{1}{m} \sum_{i=1}^{N} \| y_i-h_\theta(u_i)\|_2^2 .
\end{equation}
However, in practice sometimes only the noisy observations $y_i$ are available. In settings where the noise distribution is Gaussian, it is possible to bypass the need of clean training data with a SURE-based loss~\cite{stein1981estimation}
\begin{equation}\label{eqs:SURE_v1}
    \mathcal{L}_{\text{SURE}} (\theta)=  \sum_{i=1}^{N}  \frac{1}{m}\|y_i-h_\theta(y_i)\|^2-\sigma^2  + \frac{2\sigma^2}{m}\mathbf{\nabla} \cdot h_\theta(y_i)
\end{equation}
where $\mathbf{\nabla} \cdot h_\theta(y)=\sum_{j=1}^m\frac{\partial h_{\theta,j}(y)}{\partial y_j}$ is the divergence of the function $h_\theta$ with respect to the noisy input $y$. The first two terms in \eqref{eqs:SURE_v1} penalize the error between the noisy observations and denoised ones (bias of the denoiser), whereas the last term penalizes the variance of the denoiser. Crucially, the SURE loss is an unbiased estimator of the supervised loss:
\begin{theorem}\label{theorem:sure}\cite{stein1981estimation}
Let $h_\theta: \eR^m \rightarrow \eR^m$ be a (weakly) differentiable real-valued function such that $\forall j=1,2,\cdots,m$, $\mathbb{E}_\epsilon\{|\partial h_j(y)/\partial y_j|\}< -\infty$. Then, the SURE loss is an unbiased estimator of the supervised mean squared loss, that is
\begin{equation}
  \mathbb{E}_y \{ \mathcal{L}_{\text{SURE}} (\theta) \}=   \mathbb{E}_{y,u} \{\mathcal{L}_{\text{MSE}} (\theta)\}
\end{equation}
\end{theorem}

The SURE framework has been generalized by a number of authors, \cite{eldar2008generalized,luisier2010image,raphan2011NEBLS} beyond simple additive Gaussian noise, including Poisson and mixed Poisson Gaussian noise, resulting in different loss expressions depending on the noise type. In each case, however, there is always the need to calculate the divergence of the denoiser function (c.f.~\eqref{eqs:SURE_v1}) which may not have a simple analytic form, e.g., for deep networks. To alleviate this limitation, Ramani et al.~\cite{ramani2008monte} proposed a Monte Carlo method to estimate the divergence which only requires one additional evaluation of $h_\theta$:

\begin{theorem}\label{theorem:div}\cite{ramani2008monte}
Let $b$ be a zero-mean i.i.d. random vector with unit variance and bounded higher order moments. Then
\begin{equation}
    \mathbf{\nabla} \cdot h_\theta(y) =  \lim_{\tau\rightarrow 0}\mathbb{E}_{b}\left\{\frac{1}{\tau}b^{\top}\left(h_\theta(y+\tau b)-h_\theta(y)\right)\right\}
\end{equation}
provided that $h_\theta(y)$ admits a well-defined second-order Taylor expansion. If not, this is still valid in the weak sense provided that $h_\theta(y)$ is tempered.
\end{theorem}
Thus, the divergence term can be approximated by sampling a single i.i.d. vector, e.g., $b\sim\mathcal{N}(0,1)$, and fixing a small positive value $\tau$:
\begin{equation}\label{eqs:monte_carlo}
    \mathbf{\nabla} \cdot h_\theta(y)\approx \frac{1}{\tau}b^{\top}\left(h_\theta(y+\tau b) - h_\theta(y)\right).
\end{equation}
Incorporating this expression into the associated SURE loss has been shown to yield accurate unbiased estimates may not have a simple analytic form MSE for many various denoising methods $h_\theta(y)$ \cite{ramani2008monte}. Furthermore, combining the SURE loss \eqref{eqs:SURE_v1} with a parameterized denoiser enables unsupervised learning via minimization of this proxy for MSE without access to ground truth signals.

Recently, it has become popular to use the SURE loss for unsupervised training of denoising networks~\cite{soltanayev2018training,zhussip2019training}, and to learn solutions to compressed sensing problems where reconstruction can be cast as an iterative sequence of denoising steps, e.g.,~\cite{guo2015nopriors,metzler2016denoising,metzler2017learned,metzler2018unsupervised}. However, as explained in~\cref{learning to solve inv prob}, learning from measurement alone is not possible without further assumptions. We therefore consider next how to exploit SURE within the EI framework.

\section{Robust Equivariant Imaging}

Our aim is to develop a \emph{Robust Equivariant Imaging} (REI) framework to learn the reconstruction function $f_\theta: y\rightarrow x$ in a fully unsupervised manner from only noisy partial measurement data $\{y\}_{i=1,\dots,N}$ in inverse problems whose acquisition models are rank deficient (e.g., with a non-trivial nullspace $\nullA$ when $A$ is linear).  We assume the following generative model:
\begin{align}
    y|u \sim q_u(y), \quad
    u = A(x)
   % x \sim p(x)
\end{align}
where $q_u(y)$ is the noise distribution and  $x \sim p(x)$, the unknown signal distribution, is supported on a signal set $\mathcal{X}$ which is assumed to be invariant to the transformations  $\{T_g\}_{g=1,\dots,\ntransf}$.

\begin{figure}[t]
\begin{center}
\includegraphics[width=1\linewidth]{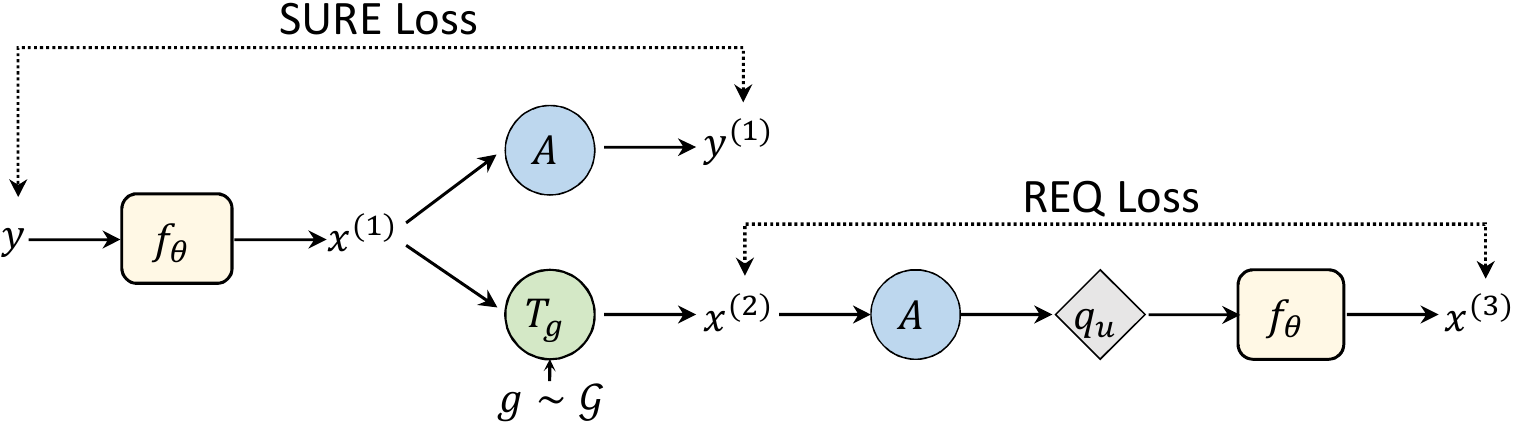}
\end{center}
\caption{\textbf{REI training strategy}. $x^{(1)}$ represents the estimated image, while $x^{(2)}$ and $x^{(3)}$  represent $T_gx^{(1)}$ and the estimate of $x^{(2)}$ from the (noisy) measurements $\tilde{y} = A (x^{(2)})$ respectively.}
\label{fig:rei}
\end{figure}

The REI model replaces the two loss terms in~\eqref{eqs:ten_loss} with noise-robust counterparts:
\begin{equation}\label{eqs:rei_loss}
    %   \mathcal{L}_{\text{REI}} =  \mathcal{L}_{\text{SURE}} + \alpha \mathcal{L}_{\text{EQ}},
      \mathcal{L}_{\text{REI}}(\theta) =  \mathcal{L}_{\text{SURE}}(\theta) + \alpha \mathcal{L}_{\text{REQ}}(\theta)
\end{equation}
The right hand term, presented in \Cref{subsec:REQ}, aims to enforce equivariance of the imaging system $f_\theta \circ A$ taking into account that the input to $f_\theta$ should be noisy. The left hand term is a SURE-based approximation of the unknown clean measurement consistency loss in~\eqref{eqs:ten_loss}, which depends on the noise model and it is presented in \Cref{subsec: sure loss}. The network can be trained in an end-to-end fashion using stochastic optimization~\cite{kingma2014adam}, see~\Cref{fig:rei}.

The proposed unsupervised training framework is agnostic of the network architecture used for  $f_\theta$. Thus, in principle, REI can be used to train any existing model in an unsupervised way. In our experiments, we use a trainable neural network $f_\theta=G_\theta\circ \eAT$, where $G_\theta: \eR^{n}\rightarrow \eR^n$ is a neural network and $\eAT:\eR^{m}\rightarrow \eR^n$ is any (fixed) suitable backprojection.

\subsection{Enforcing equivariance with noisy data} \label{subsec:REQ}

Following~\cite{chen2021equivariant}, we wish to enforce that $f_\theta \circ A$ is equivariant to transformations. We adapt the equivariant loss in~\eqref{eqs:loss_eq} to handle noisy measurements:
\begin{equation}
\mathcal{L}_{\text{REQ}}(\theta)=\sum_{i=1}^{N}\sum_{g=1}^{\ntransf}\frac{1}{\ntransf n} \|T_gf_\theta(y_i)-f_\theta(\tilde{y}_i))\|^2
\end{equation}
where $\tilde{y}_i$ is sampled from the distribution $\tilde{y}_i \sim q_{\tilde{u}_i}$ with underlying measurement $\tilde{u}_i = A\left(T_gf_\theta(y_i)\right)$. The sampled measurements $\tilde{y}_i$ can be interpreted as noisy measurements obtained via the virtual operator $A\circ T_g$. In the absence of noise, $\mathcal{L}_{\text{REQ}}(\theta)$ is equivalent to the equivariant loss in~\eqref{eqs:loss_eq}.

\subsection{SURE-based measurement consistency} \label{subsec: sure loss}
In this work, while the methodology is more general, we consider 3 specific noise models $q_u(y)$ which appear often in practice: Gaussian, Poisson and mixed Poisson-Gaussian. In each case, we develop a noise-specific measurement consistency SURE term $\mathcal{L}_{\text{SURE}}(\theta)$ which is an unbiased estimator of the unknown clean measurement consistency loss. That is we develop a SURE estimate for the denoising function $A \circ f_\theta$ that denoises the measurements:
\begin{equation}
  \mathbb{E}_y \{ \mathcal{L}_{\text{SURE}} (\theta) \}=   \mathbb{E}_{y,u} \{ \sum_{i=1}^N \frac{1}{m} \| u_i- A (f_\theta(y_i)) \|^2 \}.
\end{equation}
A detailed derivation of all the formulas is included in the supplementary material (SM).

\paragraph{Gaussian noise}
We first consider the standard Gaussian noise model, that is
\begin{equation}\label{eqs:gaussion_model}
    y \sim\mathcal{N}(u,\sigma^2I) \quad \text{with}\quad u= A(x).
\end{equation}

Following~\cref{subsec: sure}, in this case we have
\begin{equation}\label{eqs:sure_gaussian}
    \begin{split}
     &\mathcal{L}_{\text{SURE}} (\theta)=\sum_{i=1}^{N}\frac{1}{m}\|y_i - A(f_\theta(y_i))\|_2^2 -\sigma^2 \\ & +\frac{2\sigma^2}{m\tau}b_i^{\top} \left(A(f_\theta(y_i+\tau b_i)) - A(f_\theta(y_i))\right)
    \end{split}
\end{equation}
where $b_i\sim \mathcal{N}(0, I)$ and $\tau$ is a small positive number.

\paragraph{Poisson noise} In various imaging settings such as low-photon imaging, the observed measurements follow a Poisson distribution. This noise is modeled as:
\begin{equation}\label{eqs:poisson_model}
  y = \gamma z \quad \text{with}\quad z\sim \text{Poisson}(\frac{u}{\gamma}), \quad u=A(x).
\end{equation}
where $\gamma>0$ is the gain of the acquisition process which controls the noise level.
In this case, we have
\begin{equation}\label{eqs:sure_poisson}
\begin{split}
   &\mathcal{L}_{\text{SURE}}(\theta)=\frac{1}{m}\sum_{i=1}^{N}\|y_i-A(f_\theta(y_i))\|_2^2-\gamma 1^{\top}y_i\\
    &+\frac{2\gamma}{\tau}(b_i\odot y_i)^{\top} \left(A(f_\theta(y_i+\tau b_i))-A(f_\theta(y_i))\right)
\end{split}
\end{equation}
where here we select $b_i$ to be a Bernoulli random variable~\cite{le2014unbiased} taking values of $-1$ and $1$ each with a probability of $0.5$, $\tau$ is a small positive number, and $\odot$ is an elementwise multiplication.

\paragraph{Mixed Gaussian-Poisson (MPG) model} The Gaussian and Poisson noise models usually do not individually account for the various phenomena involved with real image acquisition processes in inverse problems such as fluorescence microscopy and X-ray computed tomography (CT)~\cite{le2014unbiased}. The noise in these inverse problems is better captured by a mixed Poisson-Gaussian  noise model~\cite{jezierska2011approach,le2014unbiased}:
\begin{equation}\label{eqs:mpg_model}
y = \gamma z + \epsilon \quad \text{with} \quad \left\{
\begin{array}{l}
 u = A(x) \\
z \sim \text{Poisson}\left(\frac{u}{\gamma}\right)
\\
\epsilon \sim \mathcal{N}(0, \sigma^2 I)
\end{array}
\right. .
\end{equation}
The Gaussian model is recovered as $\gamma\to 0$ whereas the Poisson model is obtained by setting $\sigma=0$.

Extending the pioneering work in \cite{luisier2010image,le2014unbiased}, we derive the unbiased risk estimator of the MSE of clean measurement consistency for the MPG noise model:
\begin{equation}\label{eqs:sure_mpg}
\begin{split}
    &\mathcal{L}_{\text{SURE}}(\theta)=\sum_{i=1}^{N}\frac{1}{m}\|y_i-A(f_\theta(y_i))\|_2^2-\gamma 1^{\top}y_i-\sigma^2\\
    &+\frac{2}{\tau}(b_i\odot (\gamma y_i + \sigma^2 I))^{\top} \left(A(f_\theta(y_i+\tau b_i))-A(f_\theta(y_i)) \right)\\
    &+\frac{2\gamma \sigma^2}{\tau}c_i^{\top} (A(f_\theta(y_i+\tau c_i)) + A(f_\theta(y_i-\tau c_i)) \ldots \\
    &- 2A(f_\theta(y_i)) )
\end{split}
\end{equation}
where $b_i\sim \mathcal{N}(0, I)$, $c_i$ are i.i.d. random variables that follow a Bernoulli distribution taking values of $-1$ and $1$ each with a probability of $0.5$, $\tau$ is a small positive number.

\section{Experiments}
We evaluated the REI  framework in 3 inverse problems: accelerated MRI, image inpainting and a (nonlinear) sparse view CT problem. More details are presented in the SM.

\begin{figure*}[t]
\begin{minipage}{1\linewidth}
\centerline{\includegraphics[width=1\textwidth]{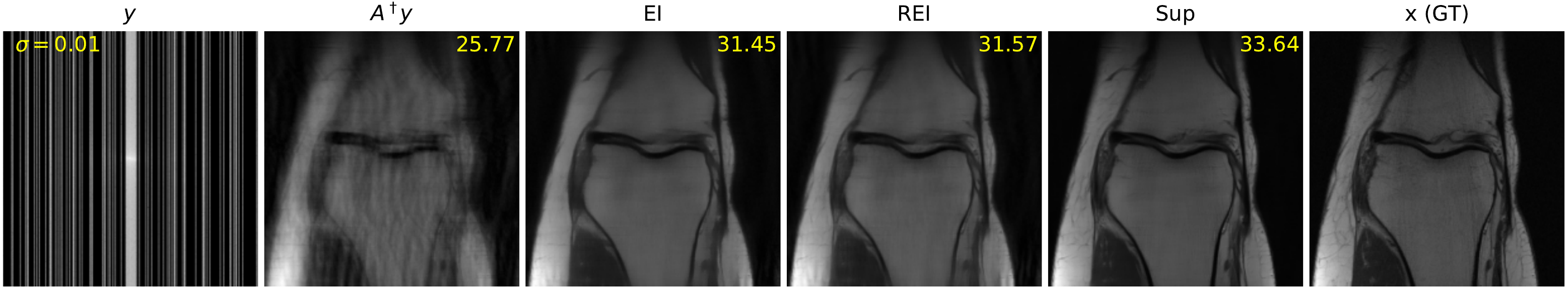}}
\end{minipage}
\begin{minipage}{1\linewidth}
\centerline{\includegraphics[width=1\textwidth]{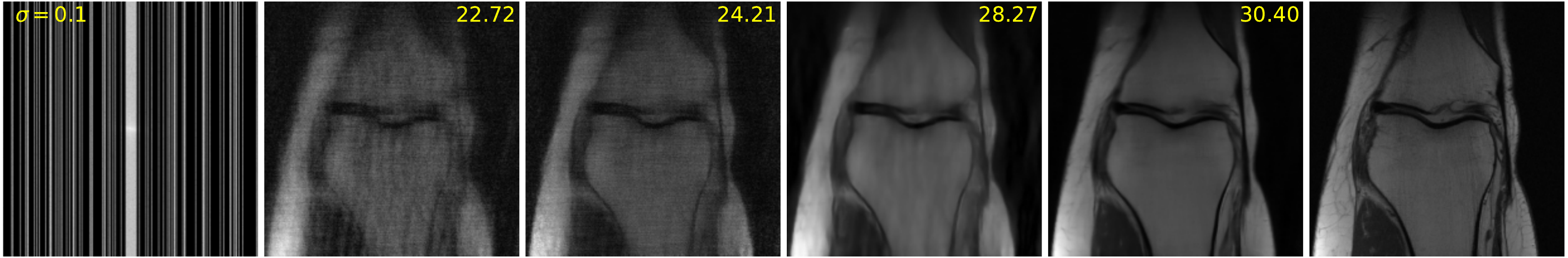}}
\end{minipage}
\begin{minipage}{1\linewidth}
\centerline{\includegraphics[width=1\textwidth]{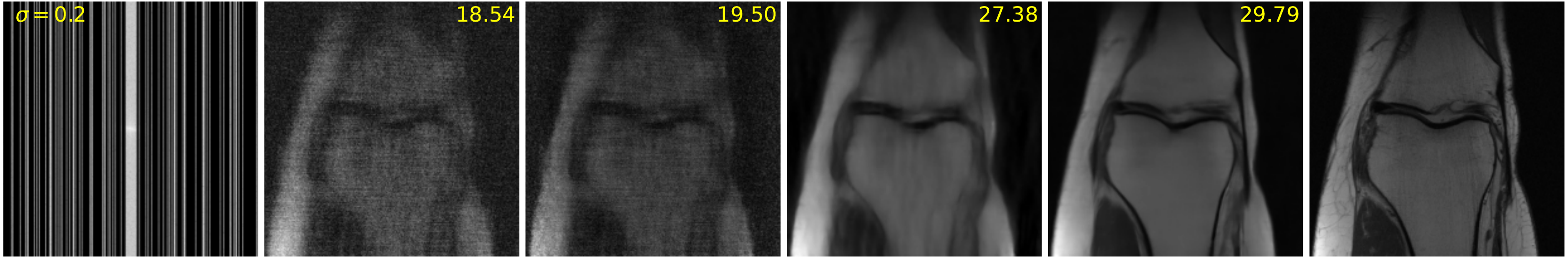}}
\end{minipage}
\caption{MRI image reconstruction on the  test data with Gaussian noise (from top to bottom: $\sigma=0.01,0.1,0.2$) and $4\times$ acceleration. PSNR values are shown in the top right corner of the images.}
\label{fig:results_mri}
\end{figure*}

\subsection{Setup and Implementation}
The forward operator $A$ in all the evaluated tasks is known and is dimension reducing ($m < n$). We considered different noise models for each task. Throughout the experiments, we follow \cite{chen2021equivariant} and use a residual U-Net~\cite{ronneberger2015u} to build $G_\theta$. We compare our new learning method (REI) with three different learning strategies: vanilla EI given by the loss in~\eqref{eqs:ten_loss}, an \emph{oracle} REI trained with noiseless measurement consistency ($\text{REI}_{\text{oracle}}$) that replaces the SURE loss by the true clean measurement consistency loss, and supervised learning (Sup)~\cite{jin2017deep}, which minimizes $\sum_{i=1}^{N}\frac{1}{n}\left\|x_i - f_\theta(y_i)\right\|_2^2$ using clean signal and noisy measurements pairs. The oracle REI provides us with a way to evaluate the separate contributions of the REQ and SURE losses. We use the same neural network architecture for all learning methods. For a fair comparison with  EI and REI,  no  data  augmentation  of  is used for  supervised  learning. We \emph{emphasize} that our goal here is not to achieve state-of-the-art performance, but to show that unsupervised learning is possible on different inverse problems with noise. The choice of architecture is somewhat orthogonal to this goal. All methods are implemented in PyTorch and optimized by Adam~\cite{kingma2014adam}.
%Please see SM for the full training details.
A full ablation study, based on the MRI experiment below, and across all the possible loss terms: EI, REI, $\text{REI}_{\text{oracle}}$, MC, SURE and combinations thereof is presented in the SM. This shows that replacement of REI or SURE with other losses has a negative impact on performance and fails to track the performance of supervised learning.

\subsection{Accelerated MRI}

MRI produces images of biological tissues by sampling the Fourier transform ($k$-space) of the image. In accelerated MRI, the goal is to recover high quality MR images from sparsely sampled $k$-space data to reduce the acquisition time. The forward operator is $A:=\mathcal{S}_{\omega}\circ\mathcal{F}$ where $\mathcal{F}$ is the 2D Fourier transform and $\mathcal{S}_{\omega}$ is a binary matrix, whose ones correspond to frequencies in $\omega$ that are measured. The pseudo-inverse $\eAT =\mathcal{F}^H \circ \mathcal{S}_{\omega}$ is the masked inverse Fourier transform.
%Usually, the sampling mask $\omega$ is known so that the forward model is fixed.
This problem has been extensively studied in the context of  compressed sensing~\cite{lustig2007sparse} and supervised deep learning~\cite{jin2017deep,hammernik2018learning,han2019k}.

The primary sources of noise in MR are electronic, and dielectric and inductive coupling to the conducting solution inside the body \cite{cardenas2008noise}, and is usually modelled as Gaussian  \cite{gudbjartsson1995rician}. We therefore train the REI models using \eqref{eqs:sure_gaussian} to implement the SURE loss in \eqref{eqs:rei_loss}. In this task, we exploit the invariance of MRI images to rotations, and use rotations of integer degree ($\ntransf$=360), while $\alpha$ is set to one (see SM for the effect of varying $\alpha$).

The MRI data used were obtained from NYU fastMRI Initiative~\cite{zbontar2018fastmri}. We trained and tested on a subset of the single-coil measurements at $4\times$ acceleration. There are $973$ samples in the dataset, the first $900$ were used for training and the last $73$ were used for testing. Each image is of size $320\times 320$ pixels. In all tests, we use complex-valued data treating the real and imaginary parts of the images as separate channels. For the purpose of visualization, we display only the magnitude images.

\begin{figure}[h]
  \begin{minipage}[c]{0.58\linewidth}
    \includegraphics[width=\linewidth]{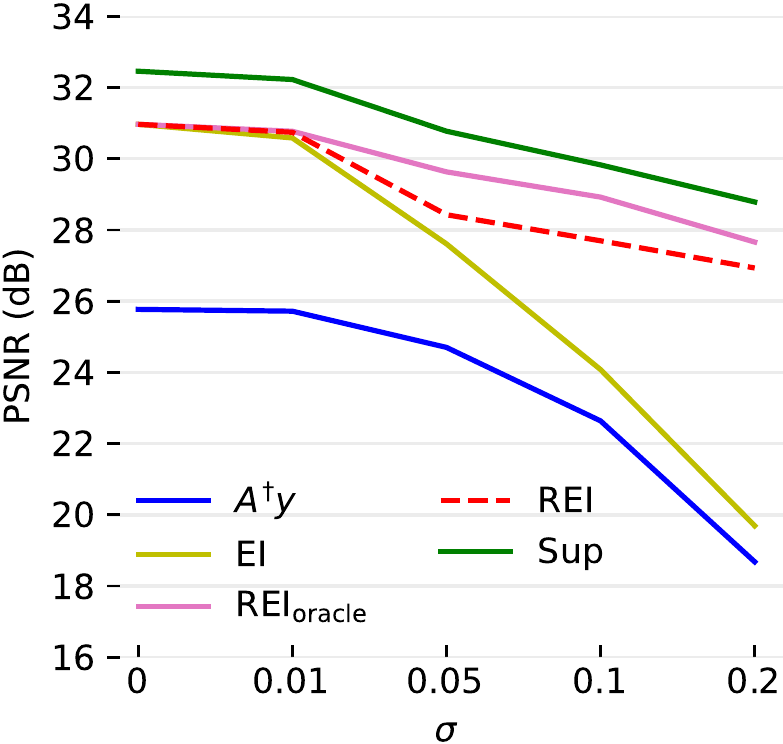}
  \end{minipage}\hfill
  \begin{minipage}[c]{0.40\linewidth}
    \caption{Reconstruction performance (PSNR) as a function of noise level $\sigma$  for linear reconstruction $\eAT (y)$, supervised training \cite{jin2017deep}, vanilla  EI \cite{chen2021equivariant}, and our method REI on  MRI image reconstruction with $4\times$ compressed and noisy $k$-space measurements.
    } \label{fig:psnr_sigma}
  \end{minipage}
\end{figure}

The reconstruction results are reported in Table \ref{table:mri} and Figures \ref{fig:results_mri}-\ref{fig:psnr_sigma}.
Note that EI only performs well for small noise levels, e.g., $\sigma \leq 0.05$, and degrades rapidly when the noise level is increased becoming almost as bad as the linear pseudo-inverse reconstruction $\eAT(y)$ for high noise levels. In comparison, REI enjoys a $3.6$ dB gain at $\sigma=0.1$ which rises to a $7$ dB improvement with respect to EI when $\sigma=0.2$. While, unsurprisingly, supervised learning always enjoys the slowest performance degradation as $\sigma$ increases due to its access to ground truth training data, the performance of REI stays within 1.5 to 2.5 dB of that of supervised learning over the full range of noise levels. The oracle REI performance lies close to that for REI, suggesting that the SURE loss~\eqref{eqs:sure_gaussian} is doing a reasonable job of approximating the (oracle) clean measurement consistency loss (see SM for the full ablation study). Overall, these results suggest that REI provides an unsupervised learning framework that is robust to noise with nearly as good performance as supervised learning.

\begin{figure*}[t]
\begin{minipage}{1\linewidth}
\centerline{\includegraphics[width=1\textwidth]{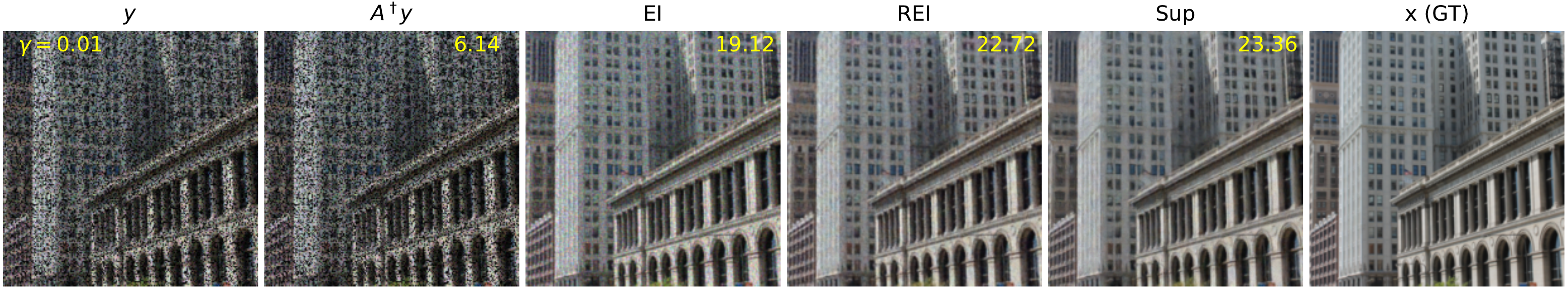}}
\end{minipage}
\begin{minipage}{1\linewidth}
\centerline{\includegraphics[width=1\textwidth]{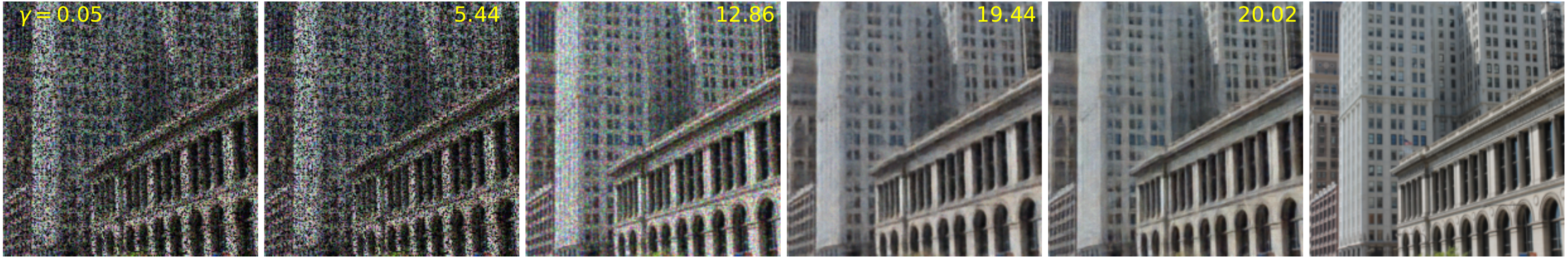}}
\end{minipage}
\begin{minipage}{1\linewidth}
\centerline{\includegraphics[width=1\textwidth]{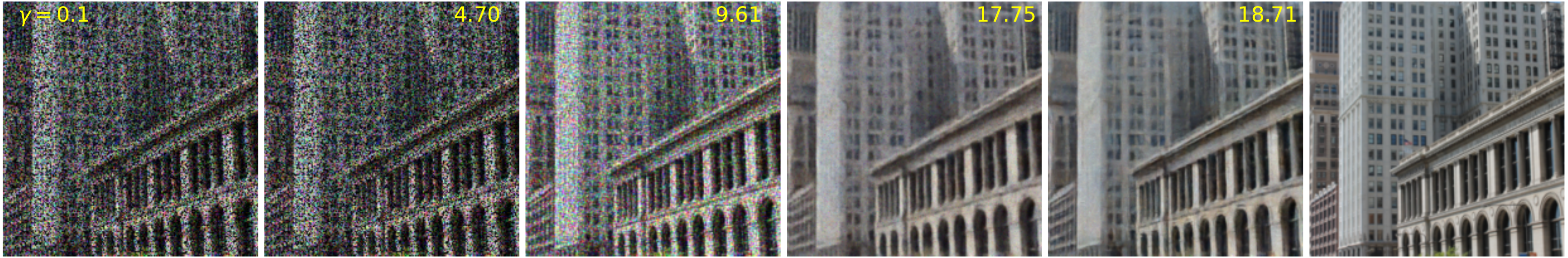}}
\end{minipage}
\caption{Inpainting reconstructions on test images with Poisson noise (from top to bottom: $\gamma=0.01,0.05,0.1$) and $30\%$ mask rate. PSNR values are shown in the top right corner of the images.}
\label{fig:results_ipt}
\end{figure*}

\subsection{Inpainting}
In image inpainting problems, the goal is to recover a complete image from a subset of noisy pixels. Inpainting can be used for example in imaging systems with hot or dead pixels. We consider the Poisson noise model in \eqref{eqs:poisson_model} to generate the noisy measurements. We therefore use \eqref{eqs:sure_poisson} to implement \eqref{eqs:rei_loss} and  to train REI models.

\begin{table}[h]
\begin{center}
\fontsize{7}{12}\selectfont
\begin{tabular}{c|ccccccccc}
$\sigma$& $\eAT (y)$&  EI  & REI & $\text{REI}_{\text{oracle}}$ & Sup\\
\hline
0 & 25.8 $\pm$ 2.7 & 31.0 $\pm$ 2.0 &31.0 $\pm$ 2.0 & 31.0 $\pm$ 2.0 &32.5 $\pm$ 2.1\\
\hline
0.01 & 25.7 $\pm$ 2.7 & 30.6 $\pm$ 1.8 & 30.8 $\pm$ 2.0 & 30.8 $\pm$ 2.0 & 32.2 $\pm$ 2.0\\
\hline
0.05 & 24.7 $\pm$ 2.1 & 27.6  $\pm$ 1.3 & 29.0  $\pm$ 1.8 & 29.6  $\pm$ 1.9  & 30.8  $\pm$ 1.8\\
\hline
0.1 & 22.6 $\pm$ 1.4 & 24.1  $\pm$ 0.9 & 27.7 $\pm$ 2.0 & 28.9  $\pm$  1.7 & 29.8  $\pm$ 1.8\\
\hline
0.2 & 18.7 $\pm$ 0.7 & 19.7 $\pm$ 0.7 & 26.9 $\pm$ 2.0 & 27.7 $\pm$ 1.7 & 28.8  $\pm$ 1.7
\end{tabular}
\end{center}
\vspace{-10pt}
\caption{PSNR of $4\times$ accelerated MRI image reconstruction for different methods using noisy test measurements.}\label{table:mri}
\end{table}

In this task, the forward model is described as $y=a\odot x$ where $a$ is a binary mask, $\odot$ is the Hadamard product, and the associated operator and pseudo-inverse are $A=\text{diag}(a)$ and $A=\eeAT$.  We consider random masks which drop $30\%$ of pixel measurements. We then train our REI models by applying random (horizontal and vertical) shift transformations with $\alpha=1$. We evaluate the reconstruction performance on the Urban100~\cite{Huang-CVPR-2015} natural image dataset. For each image, we cropped a $512\times512$ pixel area at the center and then resized it to $256\times256$ for the ground truth image. The first 90 images are used for training while the last 10 measurements are used for testing.

The results in~\Cref{table:inpainting} and Figure~\ref{fig:results_ipt} show that even for small values of $\gamma$, EI struggles to compete with REI and the performance rapidly decreases when the noise level $\gamma$ is increased. In contrast, REI performs well across all noise levels, beating EI by more than $8$ dB at high noise levels, and within $1$ dB of the fully supervised model, demonstrating the effectiveness of REI for the Poisson noise model.

%. These results are consistent with those from the previous MRI task and suggest that REI also provides a robust unsupervised learning framework for the Poisson noise model.

\begin{figure*}[t]
\begin{minipage}{1\linewidth}
\centerline{\includegraphics[width=1\textwidth]{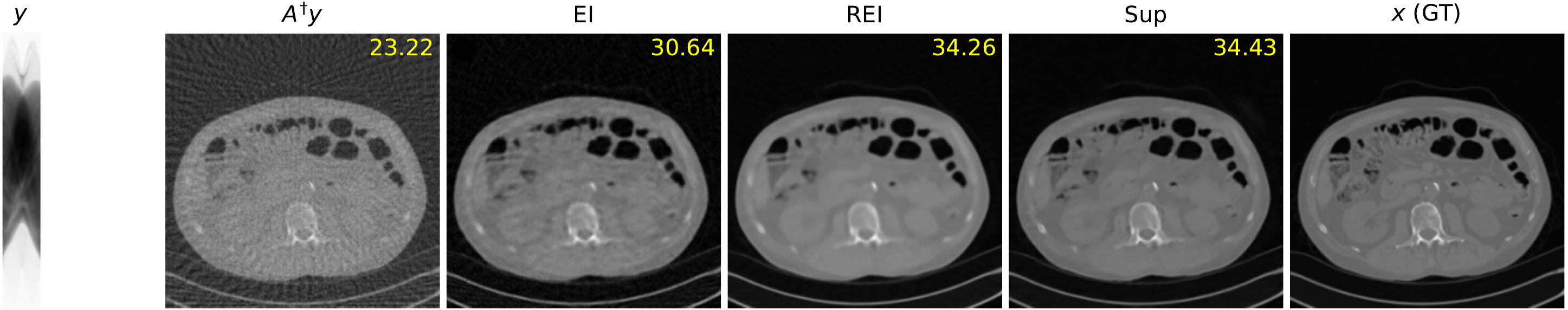}}
\end{minipage}
\begin{minipage}{1\linewidth}
\centerline{\includegraphics[width=1\textwidth]{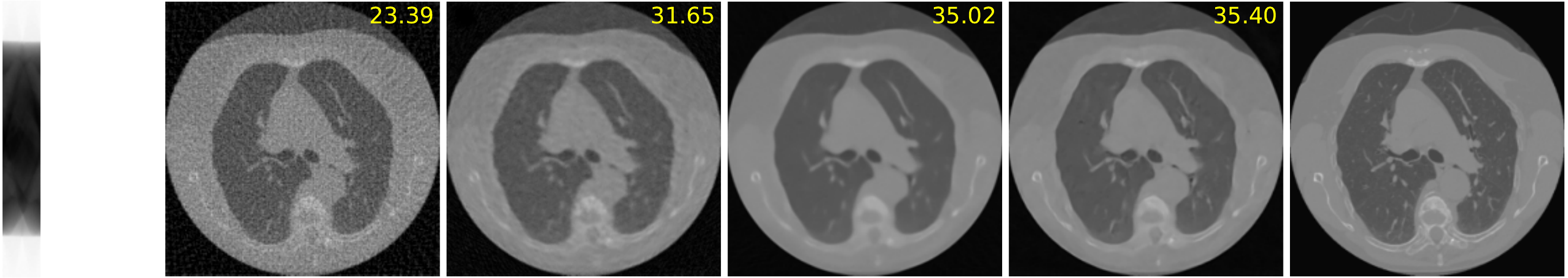}}
\end{minipage}
\caption{CT image reconstruction (50 views) on the  test observations with mixed Poisson-Gaussian noise, $I_0=10^5$, $\sigma=30$, $\gamma=1$. PSNR values are shown in the top right corner of the images.}
\label{fig:results_ct}
\end{figure*}
%\vspace{-8pt}

\begin{table}[h]
\begin{center}
\fontsize{7}{12}\selectfont
\begin{tabular}{c|ccccccccc}
$\gamma$& $\eAT (y)$&  EI & REI & $\text{REI}_{\text{oracle}}$  & Sup\\
\hline
0.01 & 5.7 $\pm$ 1.5& 18.9  $\pm$ 1.0 & 21.7 $\pm$ 1.3 & 22.0  $\pm$ 1.3 & 22.3 $\pm$ 1.1\\
\hline
0.05 & 5.1 $\pm$ 1.4 & 11.8 $\pm$ 3.0 & 18.2 $\pm$ 1.2 & 18.3 $\pm$ 0.9 & 19.6 $\pm$ 1.2\\
\hline
0.1 &  4.4 $\pm$ 1.3 & 9.8 $\pm$ 0.8 & 16.6 $\pm$ 1.3  & 17.3 $\pm$ 1.3  & 18.2 $\pm$ 1.2
\end{tabular}
\end{center}
\vspace{-10pt}
\caption{PSNR of pixelwise image inpainting reconstruction for different methods using noisy test measurements.}\label{table:inpainting}
\end{table}

\subsection{Sparse-view CT}
Our final experiment provides an initial demonstration of robust equivariant imaging for a nonlinear inverse problem. Sparse-view CT can be described by a nonlinear forward model $A(x)=I_0e^{-\texttt{radon}(x)}$, where \texttt{radon} denotes a sparse-view Radon transformation and $I_0$ denotes the X-ray source intensity.
Quantum and electronic noise are two major noise sources in X-ray CT scanners~\cite{ding2018statistical} and for normal clinical exposures, measurements $y$ are often modeled as the sum of a Poisson distribution representing photon-counting statistics and an independent Gaussian distribution representing additive electronic noise, i.e., $y = z + \epsilon$ where $z \sim \text{Poisson}(A(x))$ and $\epsilon \sim \mathcal{N}(0, \sigma^2 I)$.

Here, we exploit the invariance of CT images to rotations and consider rotations of multiples of $1$ degree ($\ntransf$=360).
For $\eeAT$ we use the filtered backprojection (FBP), \texttt{iradon}, of the logarithm of the measurements, $\log(I_0/y)$.
 The reconstruction function is thus defined as $f_\theta(y) = G_\theta\circ \texttt{iradon}\circ \log(I_0/y)$. We therefore use the MPG model, \eqref{eqs:sure_mpg} to implement \eqref{eqs:rei_loss} and train the REI models.

We use the CT100 clinical dataset \cite{clark2013cancer}, which comprises 100 real in-vivo CT images collected from the cancer imaging archive.  The CT images were resized to $256\times 256$ pixels. The \texttt{radon} function consists of 50 uniformly sampled views (angles). For the imaging noise we follow \cite{ding2018statistical} and generate MPG noisy measurements with $I_0=10^5$, $\sigma=30$ and $\gamma=1$. We used the sinograms of the first 90 images for training and the remaining 10 for testing. In this task, we use the U-Net which is the same architecture as FBPConvnet in~\cite{jin2017deep}. We trained EI, REI and $\text{REI}_{\text{oracle}}$ with $\alpha=10^3$ (this is much larger than in the MRI and inpainting experiments due to the scaled log, $\log(I_0/y)$)\footnote{Generally, $\alpha$ should scale inversely proportional to the norm of the (linearized) measurement operator.} using the sinograms $y$ alone while the supervised training is performed with the noisy measurements and ground truth pairs $(x,y)$.

A quantitative comparison is presented in Table~\ref{table:ct}. While all learning-based methods outperform the standard FBP reconstruction, REI achieves a $5$ dB gain when compared with EI, and performs almost as well as the oracle $\text{REI}_{\text{oracle}}$ and the fully supervised baseline. Figure~\ref{fig:results_ct} shows examples of reconstructed images. The FBP reconstruction contains massive line artefacts and noise due to the sparse-view MPG measurements. Although EI performs better, it  fails to fully remove the artefacts and noise. In contrast, REI achieves almost equivalent quality to the supervised method. A similarly performing low-dose example, with $I_0=10^4$, $\sigma=50$, is given in the SM.

\begin{table}[h]
\begin{center}
\fontsize{7}{12}\selectfont
\begin{tabular}{c|ccccc}
&FBP&  EI & REI & $\text{REI}_{\text{oracle}}$ &Sup\\
\hline
CT &22.8 $\pm$ 0.5& 29.2 $\pm$ 0.6 & 34.0 $\pm$ 1.1 & 34.0 $\pm$ 1.2 & 34.4 $\pm$ 1.1
\end{tabular}
\end{center}
\vspace{-10pt}% only for non-anonymous
\caption{Reconstruction performance (PSNR) of sparse-view CT task for different methods on the test noisy measurements.}\label{table:ct}
\end{table}
\vspace{-8pt}

\section{Discussion}
One might naturally ask whether one could hard wire the equivariance into the neural network. Indeed, there has been considerable efforts recently in designing such architectures,
%that are equivariant to group actions,
mostly in the context of classification~\cite{cohen2016group,kondor2018compactgroups,mallat2012group}. However, in equivariant imaging it is not the network that we want to be equivariant but the full acquisition-reconstruction system, $f_\theta \circ A$. Celledoni et al.~\cite{celledoni2021equivariant} proposed an interesting extension to design equivariant networks for inverse problems by using an unrolled network with an equivariant network substituting the proximal component. However, as noted by the authors, this approach does not guarantee full system equivariance and is restricted to supervised learning applications. In contrast, (R)EI promotes full system equivariance through the (robust) equivariant loss, thereby enabling unsupervised learning.

Training using SURE requires knowledge of the noise characteristics.\footnote{Robustness to different noise levels can be achieved by training over different levels, or possibly by incorporating ideas from~\cite{mohan2020}.} While we evaluated REI on the Gaussian, Poisson and MPG models, SURE can handle many other models including non-exponential ones, see \cite{raphan2011NEBLS} for a detailed list. SURE based training losses have also been used in other related works,~\cite{kim2020unsupervised,zhussip2019extending,metzler2018unsupervised,soltanayev2018training} for unsupervised imaging/denoising, however they are only applied to inverse problems with complete measurements.  %(i.e., a trivial nullspace if $A$ is linear).
Noise2Noise~\cite{lehtinen2018noise2noise}
%, a special case of extended SURE \cite{zhussip2019extending},
takes a slightly different route (and only considered denoising) by using training pairs consisting of two different noisy realisations of the same signal. While this may appear to be blind to the noise characteristics, the requirement to have two distinct noisy realisations implicitly provides such information. Indeed, we emphasise that some such prior information is always required in any fully unsupervised denoising technique.

Finally, some papers have proposed unsupervised  methods for compressed sensing (with non-trivial forward operators)~\cite{guo2015nopriors,metzler2018unsupervised}. However, these methods rely heavily on the randomness of $A$. In contrast, REI applies a SURE-based loss for measurement consistency while exploiting equivariance to encourage a unique reconstruction from partial measurements.

\section{Conclusion}
In this paper, we proposed {\em Robust Equivariant Imaging}, a fully unsupervised imaging method that can effectively learn to image from only noisy partial measurements. The method only requires knowledge about the (possibly nonlinear) forward operator and the acquisition noise characteristics, which for most imaging systems are usually known. The framework is architecture agnostic, and can be used to train any existing model, and a wide range of noise models in a fully unsupervised way.
%Furthermore, it can handle a wide range of noise models including but not limited to the ones considered in this paper \dc{(see table 1 in \cite{raphan2011NEBLS} for a non-exhautive list of SURE applicable models)}.

\section*{Acknowledgments}
This work is supported by the ERC C-SENSE project (ERCADG-2015-694888).

%%%%%%%%% REFERENCES
{\small
\bibliographystyle{ieee_fullname}
\bibliography{egbib}
}

% \newpage
% \clearpage

\clearpage
\newpage
\appendix

\section*{Appendix}
\section{SURE-based losses}
We derive the SURE loss presented in the main paper for the Gaussian, Poisson and Poisson-Gaussian noise models.
We follow the derivation in~\cite{le2014unbiased}. In all cases, the goal is to obtain an unbiased estimator of the supervised mean squared error (MSE) of the clean measurement $u$ from the noisy measurement $y$:
\begin{equation} \label{eq: MSE}
   \sum_{i=1}^{N} \frac{1}{m}\| u_i - h_\theta (y_i) \|^2
\end{equation}
with denoiser $h_\theta:\eR^{m}\mapsto\eR^{m}$ defined as
\begin{equation}
    h_\theta = A \circ f_\theta
\end{equation}
where $A:\eR^{n}\mapsto\eR^{m}$ denotes the forward operator and $f_\theta:\eR^{m}\mapsto\eR^{n}$ is the (trainable) reconstruction network. The expectation of~\eqref{eq: MSE} with respect to the pairs $(y,u)$ can be decomposed as
\begin{equation}
\begin{split}
  \mathbb{E}_{y,u}& \{\sum_{i=1}^{N} \frac{1}{m}\| u_i - h_\theta (y_i) \|^2 \} \\
  &= \mathbb{E}_{u} \sum_{i=1}^{N} \frac{1}{m} \mathbb{E}_{y|u}\| u_i - h_\theta (y_i) \|^2.
\end{split}
\end{equation}
The inner expectation can be further decomposed as
\begin{align*}
&\mathbb{E}_{y|u}\| u_i - h_\theta (y_i) \|^2  \\
  %&= \mathbb{E}_{u} \mathbb{E}_{y|u} \| u_i - h_\theta (y_i) \|^2  \\
 &=  \mathbb{E}_{y|u} \{\| u_i \|^2 + \| h_\theta(y_i)\|^2  - 2u_i^{\top}h_\theta(y_i) \} \\
  &= \mathbb{E}_{y|u} \{ u_i^{\top} y_i \} + \mathbb{E}_{y|u} \{\| h_\theta(y_i)\|^2 \}  - 2  \mathbb{E}_{y|u}\{ u_i^{\top}h_\theta(y_i) \}
\end{align*}
where we used that $\mathbb{E}_{y|u}\{y_i\}=u_i$ for all noise models.
An unbiased estimator of the second term is simply $\| h_\theta(y_i)\|^2$, which does not require clean measurements $u$. The terms
\begin{equation} \label{eq:diffucult term}
    \mathbb{E}_{y|u} \{u_i^{\top}h_\theta(y_i) \}
\end{equation}
and
\begin{equation} \label{eq:diffucult term id}
    \mathbb{E}_{y|u} \{u_i^{\top}y_i \}
\end{equation}
depend on $u$ and require a noise-dependent analysis, which is presented in the following subsections.

\subsection{Gaussian}
We begin with the Gaussian noise model. In this case, we have $y \sim \mathcal{N} (u, \sigma^2 I)$ where $I$ is an $m\times m$ identity matrix. The terms~\eqref{eq:diffucult term} and~\eqref{eq:diffucult term id} can be computed in an unsupervised way (without clean $u$) using the following lemma:

\begin{lemma}[Lemma 2 in~\cite{stein1981estimation}] \label{lemma:gaussian}
Let $y\in\eR^{m}$ such that $y \sim \mathcal{N} (u, \sigma^2 I)$ be a random variable and let $\phi: \eR^{m}\mapsto\eR^{m}$ be a weakly differentiable function such that $\mathbb{E}_{y|u}  \{ | \delta \phi_{j}(y)/\delta y_j | \} <\infty$ for all $j$ and input $y$. Then,
\begin{equation}
    \mathbb{E}_{y|u} \{u^{\top}\phi(y) \} = \mathbb{E}_{y|u}\{ y^{\top} \phi(y) -  \sigma^2 \nabla \cdot \phi(y)\}
\end{equation}
where $ \nabla \cdot \phi(y) = \sum_j \frac{\delta \phi_j(y)}{\delta y_j}$ denotes the divergence of $\phi$.
\end{lemma}

Applying \Cref{lemma:gaussian} to~\eqref{eq:diffucult term} with $\phi = h_\theta$ and to~\eqref{eq:diffucult term id} with $\phi=$ identity, we get the following unbiased estimator of the MSE:

\begin{multline*}
 \sum_{i=1}^{N}  \frac{1}{m} \{\| y_i\|^2 - \sigma^2 m +  \| h_\theta(y_i)\|^2  -2 y^{\top} h_\theta(y) +\\+ 2 \sigma^2 \nabla \cdot h_\theta(y_i) \}
\end{multline*}
\begin{align*}
     = \sum_{i=1}^{N}  \frac{1}{m} \| y_i - h(y_i)\|^2  - \sigma^2+ \frac{2 \sigma^2}{m} 1^{\top} \delta h_\theta(y_i)
\end{align*}
where $\delta h_\theta(y) = [\frac{\delta h_\theta(y)}{\delta y_1},\dots, \frac{\delta h_\theta(y)}{\delta y_m}]^{\top}$ is the gradient of $h_\theta$ with respect to $y$.

Using a Monte Carlo approximation of the last term (c.f., Theorem 2 of the main paper), and decomposing $h_\theta = A \circ f_\theta$, we obtain the unsupervised loss used in the main paper:
\begin{equation}\label{eqs:sure_gaussian_sm}
    \begin{split}
     &\mathcal{L}_{\text{SURE}} (\theta)=\sum_{i=1}^{N}\frac{1}{m}\|y_i - A(f_\theta(y_i))\|_2^2 -\sigma^2 \\ & +\frac{2\sigma^2}{m\tau}b_i^{\top} \left(A(f_\theta(y_i+\tau b_i)) - A(f_\theta(y_i))\right)
    \end{split}
\end{equation}
where $b_i\sim \mathcal{N}(0, I)$ and $\tau$ is a small positive number.
\subsection{Poisson}
In the Poisson noise case, the noisy measurements are modeled as $y = \gamma z$ with $z\sim\text{Poisson}(\frac{u}{\gamma})$. The following lemma provides unsupervised expressions for~\eqref{eq:diffucult term} and~\eqref{eq:diffucult term id}.
\begin{lemma}[Lemma 1.2 in~\cite{le2014unbiased}]\label{lemma:poisson}
Let $z\in\eR^{m}$ such that $z\sim\text{Poisson} (u)$ be a random variable and let $\phi: \eR^{m}\mapsto\eR^{m}$ be a function such that $\mathbb{E}_{z|u}  \{ | \phi_{j}(z) | \} <\infty$ for all $j$.
\begin{equation}
    \mathbb{E}_{z|u} \{u^{\top}\phi(z) \} =  \mathbb{E}_{z|u}\{ z^{\top} \phi^{[-1]}(z)\}
\end{equation}
where the $j$th entry of the vector $\phi^{[-\alpha]}(z)$ is given by $\phi_{j}(y-\alpha e_j)$ and $e_j$ is the $j$th canonical vector.
\end{lemma}
Let $\phi(z) = h_\theta(y) = h_\theta(\gamma z)$, then term~\eqref{eq:diffucult term} is given by
\begin{align}
    \mathbb{E}_{y|u} \{u_i^{\top}h_\theta(y_i) \} &=  \gamma \mathbb{E}_{z|u} \{(\frac{u_i}{\gamma})^{\top}\phi(z_i) \} \\
    &= \gamma \mathbb{E}_{z|u}\{ z_i^{\top} \phi^{[-1]}(z_i)\}  \\
    &=  \mathbb{E}_{y|u}\{ y_i^{\top} h_\theta^{[-\gamma]}(y_i)\}
\end{align}
and an unbiased estimator is simply
\begin{equation}
    y_i^{\top} h_\theta^{[-\gamma]}(y_i)
\end{equation}
Now let $\phi(z) = y = \gamma z$ be the identity function, term~\eqref{eq:diffucult term id} is given by
\begin{align}
    \mathbb{E}_{y|u} \{u_i^{\top}y_i \} &= % \gamma \mathbb{E}_{z|u} \{(\frac{u_i}{\gamma})^{\top}\phi(z_i) \} \\
%    &= \gamma \mathbb{E}_{z|u}\{ z_i^{\top} \phi^{[-1]}(z_i)\}  \\
      \mathbb{E}_{y|u}\{ y_i^{\top} (y_i-1\gamma)\}
\end{align}
where $1$ denotes a vector of $m$ ones. An unbiased estimator is given by
\begin{equation}
    y_i^{\top} (y_i-1\gamma).
\end{equation}

Thus, an unbiased estimator of the MSE is given by
\begin{align*}
    &\sum_{i=1}^{N} \frac{1}{m}\{  \| y_i \|^2 - \gamma1^{\top}y_i  +  \| h_\theta(y_i)\|^2   - 2  y_i^{\top} h_\theta^{[-\gamma]}(y_i)  \}\\
    & \approx \sum_{i=1}^{N}  \frac{1}{m}\{  \| y_i \|^2 - \gamma1^{\top}y_i  +  \| h_\theta(y_i)\|^2   + 2\gamma  y_i^{\top}  \delta h_\theta(y_i) \} \\
    & = \sum_{i=1}^{N}  \frac{1}{m} \| y_i - h_\theta(y_i)\|^2 -  \frac{\gamma}{m}1^{\top}y_i +  \frac{2\gamma}{m}  y_i^{\top}  \delta h_\theta(y_i)
\end{align*}
where we used a Taylor expansion to approximate $h_\theta^{[-\gamma]}(y)\approx h_\theta(y) - \gamma \delta h_\theta(y) $~\cite{le2014unbiased}. Using a Monte Carlo estimate of the last term~\cite{le2014unbiased} (similar to Theorem 2 in the main paper) and decomposing $h_\theta = A \circ f_\theta$, we get the unsupervised loss of the main paper
\begin{equation}\label{eqs:sure_poisson_sm}
\begin{split}
   &\mathcal{L}_{\text{SURE}}(\theta)=\sum_{i=1}^{N}\frac{1}{m}\|y_i-A(f_\theta(y_i))\|_2^2-\frac{\gamma}{m} 1^{\top}y_i\\
    &+\frac{2\gamma}{m\tau}(b_i\odot y_i)^{\top} \left(A(f_\theta(y_i+\tau b_i))-A(f_\theta(y_i))\right)
\end{split}
\end{equation}
where $b$ is an i.i.d. random vector following a Bernoulli distribution.

\subsection{Poisson-Gaussian}
Noisy measurements under a Poisson-Gaussian noise model are defined by
\begin{equation}\label{eq:mpg_model}
y = \gamma z + \epsilon \quad \text{with} \quad \left\{
\begin{array}{l}
 u = A(x) \\
z \sim \text{Poisson}\left(\frac{u}{\gamma}\right)
\\
\epsilon \sim \mathcal{N}(0, \sigma^2 I)
\end{array}
\right.
\end{equation}
where $\gamma>0$ controls the Poisson noise contribution and $\sigma>0$ controls the Gaussian noise contribution.
%The Gaussian model is recovered as $\gamma\to 0$ whereas the Poisson model is obtained by setting $\sigma=0$.
An unsupervised equivalent of~\eqref{eq:diffucult term} and~\eqref{eq:diffucult term id} can be obtained using \Cref{lemma:gaussian,lemma:poisson}.
Let $\phi_\epsilon(z) = h_\theta(y) = h_\theta(\gamma z + \epsilon)$, then \eqref{eq:diffucult term} is given by
\begin{align}
    \mathbb{E}_{y|u} &\{u_i^{\top}h_\theta(y_i) \} =  \gamma \mathbb{E}_{\epsilon|u} \mathbb{E}_{z|u} \{(\frac{u_i}{\gamma})^{\top}\phi_\epsilon(z_i) \} \\
    &= \gamma \mathbb{E}_{y|u}\{ z_i^{\top} \phi_\epsilon^{[-1]}(z_i)\} \label{eq:poisstrick} \\
    &=  \gamma \mathbb{E}_{y|u}\{ \frac{y_i-\epsilon}{\gamma}^{\top} h_\theta^{[-\gamma]}(y_i)\} \\
    &=   \mathbb{E}_{y|u}\{ y_i^{\top} h_\theta^{[-\gamma]}(y_i)\} - \mathbb{E}_{y|u}\{ \epsilon_i^{\top} h_\theta^{[-\gamma]}(y_i)\}  \\
    &= \mathbb{E}_{y|u}\{ y_i^{\top} h_\theta^{[-\gamma]}(y_i) - \sigma^2 \nabla \cdot h_\theta^{[-\gamma]}(y_i)\} \label{eq:gausstrick}
\end{align}
where we used \Cref{lemma:poisson} in~\eqref{eq:poisstrick} and \Cref{lemma:gaussian} in~\eqref{eq:gausstrick}. Thus,
\begin{align}
   y_i^{\top} h_\theta^{[-\gamma]}(y_i) - \sigma^2 \nabla \cdot h_\theta^{[-\gamma]}(y_i)
\end{align}
 is an unbiased estimator of \eqref{eq:diffucult term}.
Similarly, setting $\phi_\epsilon(z) = y = \gamma z+ \epsilon$, \eqref{eq:diffucult term id} is  equal to
\begin{align}
    \mathbb{E}_{y|u} &\{u_i^{\top}y_i \} = \mathbb{E}_{y|u}\{ y_i^{\top} (y_i-1\gamma) - \sigma^2 m\} %\label{eq:gausstrick}
\end{align}
 and
 \begin{align}
   y_i^{\top} (y_i-1\gamma) - \sigma^2 m
\end{align}
is an unbiased estimator of \eqref{eq:diffucult term id}.
Thus, an unbiased estimator of the MSE is given by
\begin{multline*}
    \sum_{i=1}^{N} \frac{1}{m}\{  \| y_i \|^2 - \gamma1^{\top}y_i - \sigma^2m +  \| h_\theta(y_i)\|^2   -  \\ -2  y_i^{\top} h_\theta^{[-\gamma]}(y_i) + 2 \sigma^2 \nabla \cdot h_\theta^{[-\gamma]}(y_i)   \}
\end{multline*}
\begin{multline*}
     \approx  \sum_{i=1}^{N}  \frac{1}{m} \| y_i - h_\theta(y_i)\|^2 -  \frac{\gamma}{m}1^{\top}y_i - \sigma^2 ´+   \\ + \frac{2}{m} (\gamma y_i^{\top}  +  \sigma^21^{\top}) \delta h_\theta(y) - \frac{2\sigma^2 \gamma}{m} 1^{\top} \delta^2 h_\theta (y)
\end{multline*}
where $\delta^2 h_\theta(y) = [\frac{\delta^2 h_\theta(y)}{\delta y_1^2},\dots, \frac{\delta^2 h_\theta(y)}{\delta y_m^2}]^{\top}$ is a vector of second derivatives of $h_\theta$ with respect to $y$.
Using a Monte Carlo approximation of the last two terms~\cite{le2014unbiased} (similar to Theorem 2 in the main paper) and decomposing $h_\theta = A \circ f_\theta$, we obtain the unsupervised loss in the main paper:
\begin{equation}\label{eqs:sure_mpg_sm}
\begin{split}
    &\mathcal{L}_{\text{SURE}}(\theta)=\sum_{i=1}^{N}\frac{1}{m}\|y_i-A(f_\theta(y_i))\|_2^2-\frac{\gamma}{m} 1^{\top}y_i-\sigma^2\\
    &+\frac{2}{m\tau}(b_i\odot (\gamma y_i + \sigma^2 I))^{\top} \left(A(f_\theta(y_i+\tau b_i))-A(f_\theta(y_i)) \right)\\
    &+\frac{2\gamma \sigma^2}{m\tau}c_i^{\top} (A(f_\theta(y_i+\tau c_i)) + A(f_\theta(y_i-\tau c_i)) \ldots \\
    &- 2A(f_\theta(y_i)) )
\end{split}
\end{equation}
where $b_i\sim \mathcal{N}(0, I)$, $c_i$ are i.i.d. random variables that follow a Bernoulli distribution.

\section{Training Details} \label{sec:details}
\begin{figure}[t]
\begin{center}
\includegraphics[width=1\linewidth]{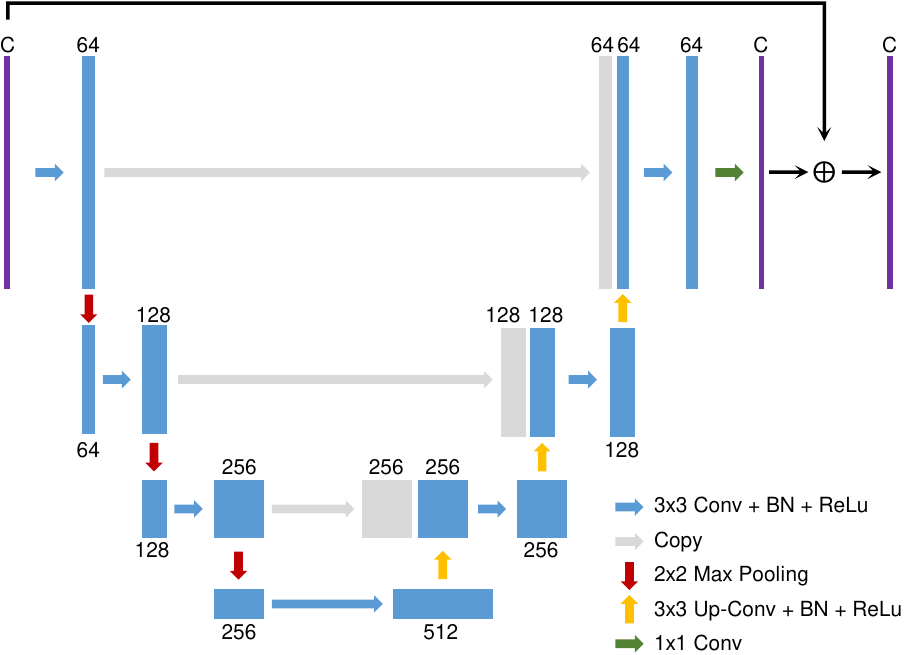}
\end{center}
\caption{The residual U-Net \cite{ronneberger2015u} used in the paper. The number of input and output channels is denoted as $C$, with $C=1, 2, 3$ in the CT, MRI and inpainting task, respectively.}
\label{fig:unet}
\end{figure}

We first provide the details of the network architectures and hyperparameters used in simulations for Figures 1, 4-7 and Tables 1-3 of the main paper. We implemented the algorithms and operators (e.g.,~\texttt{radon} and \texttt{iradon}) in Python with PyTorch 1.6 and trained the models on NVIDIA 1080ti and 2080ti GPUs. We used the  network architecture in~\cite{chen2021equivariant} for defining $G_\theta$. Figure \ref{fig:unet} illustrates the architecture of the residual U-Net~\cite{ronneberger2015u} used  in our paper. The training details for each task are as follows:

\paragraph{Accelerated MRI.} For the $4\times$ accelerated MRI task, we used Adam with a batch size of 2 and an initial learning rate of $5\times 10^{-4}$. The weight decay is $10^{-8}$. We trained the networks for $500$ epochs,  keeping the learning rate constant for the first 300 epochs and then shrinking it by a factor of $0.1$. We set $\alpha=1$ and $\tau=0.01$.

\paragraph{Inpainting.} For the inpainting task, we also used Adam but with a batch size of 1 and an initial learning rate of $10^{-4}$. The weight decay is $10^{-8}$. We trained the networks for $500$ epochs,  shrinking the learning rate by a factor of $0.1$ every 100 epochs. We set $\alpha=1$ and $\tau=0.01$.

\paragraph{Sparse-view CT.} For the $50$-views CT task, we used the Adam optimizer with a batch size of $2$ and an initial learning rate of $5\times 10^{-4}$. The weight decay is $10^{-8}$.  We trained the networks for $3000$ epochs, shrinking the learning rate by a factor of $0.1$ every 1000 epochs.
% keeping the learning rate constant for the first 2000 epochs and then shrinking it by a factor of $0.1$.
In particular,  because the CT model is nonlinear and involves an exponential mapping such that the consistency loss for $y$ is very large, we  scale the $\mathcal{L}_{\text{SURE}}$ (in Equation~(14) of main paper) accordingly by a factor of $10^{-5}$. We set $\alpha=1000$ and $\tau=10$.

\begin{figure}[t]
%   \begin{minipage}[c]{0.70\linewidth}
    \includegraphics[width=\linewidth]{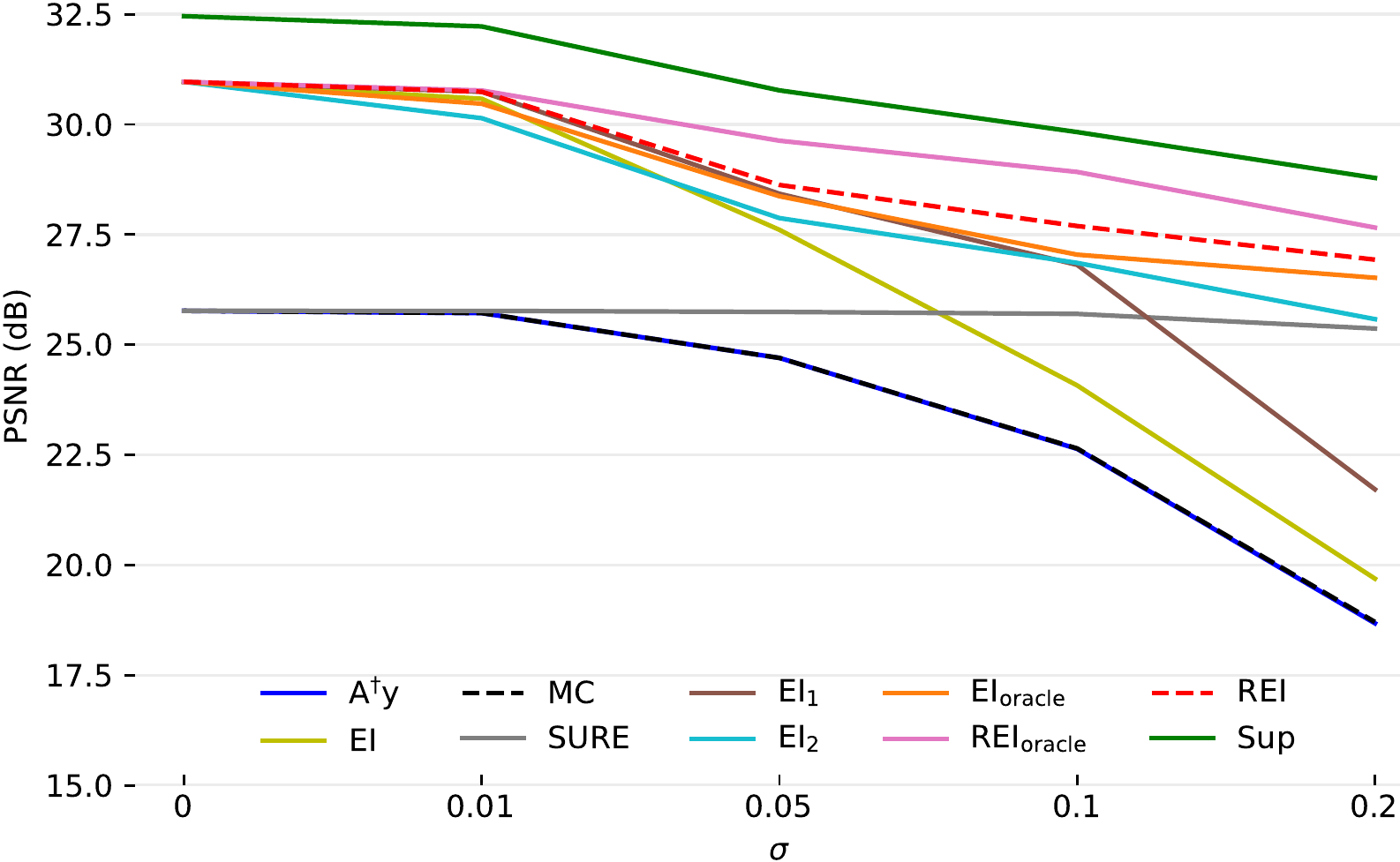}
%   \end{minipage}\hfill
%   \begin{minipage}[c]{0.30\linewidth}
    \caption{Reconstruction performance (PSNR) as a function of noise level $\sigma$  for different training losses on the MRI image reconstruction task with $4\times$ compression and noisy $k$-space measurements.}
    \label{fig:ei_variants}
%   \end{minipage}
\end{figure}

\begin{figure*}[t]
\begin{minipage}{1\linewidth}
\centerline{\includegraphics[width=1\textwidth]{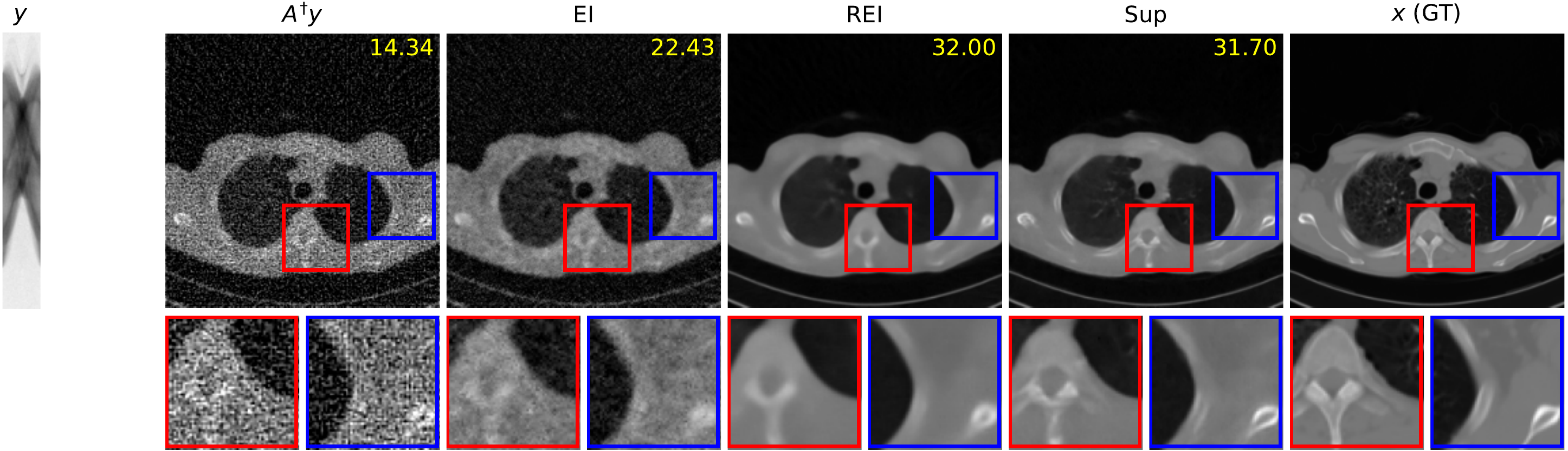}}
\end{minipage}
\begin{minipage}{1\linewidth}
\centerline{\includegraphics[width=1\textwidth]{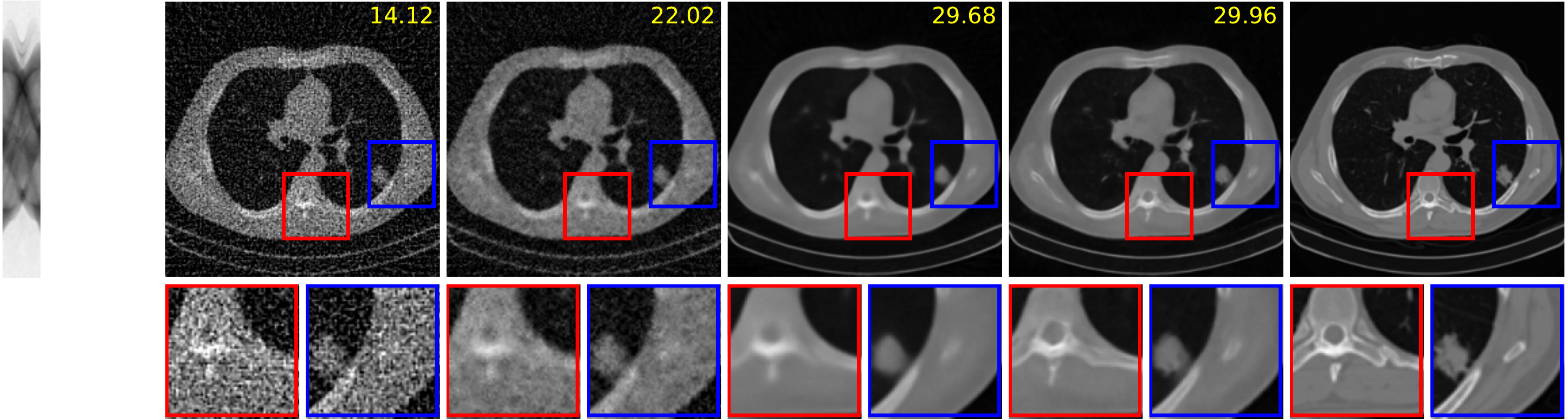}}
\end{minipage}
\caption{\emph{Low-dose} CT image reconstruction (50 views) on the  test observations with mixed Poisson-Gaussian noise, $I_0=10^4$, $\sigma=50$, $\gamma=1$. PSNR values are shown in the top right corner of the images.}
\label{fig:results_ct_sm}
\end{figure*}

\section{More results}
\subsection{Ablation study on training loss} We first compared different variants of loss function for training the REI models. In particular, we consider the following $9$ training loss variants:

\begin{itemize}
    \item \textbf{MC}: $\mathcal{L}_{\text{MC}}(\theta)$
    \item \textbf{SURE}:  $\mathcal{L}_{\text{SURE}}(\theta)$
    \item \textbf{EI}: $\mathcal{L}_{\text{MC}}(\theta)$ + $\alpha\mathcal{L}_{\text{EQ}}(\theta)$
    % \item \textbf{$\text{EI}_{\text{REQ}}$}: $\sum_{i=1}^{N}\frac{1}{m}\|y_i-Af_\theta(y_i)\|^2 + \alpha\mathcal{L}_{\text{REQ}}(\theta)$
        \item \textbf{$\text{EI}_{\text{1}}$}: $\mathcal{L}_{\text{MC}}(\theta) + \alpha\mathcal{L}_{\text{REQ}}(\theta)$
    % \item \textbf{$\text{EI}_{\text{SURE}}$}: $\mathcal{L}_{\text{SURE}}(\theta) +\alpha \mathcal{L}_{\text{EQ}}(\theta)$
        \item \textbf{$\text{EI}_{\text{2}}$}: $\mathcal{L}_{\text{SURE}}(\theta) +\alpha \mathcal{L}_{\text{EQ}}(\theta)$

    \item \textbf{$\text{EI}_{\text{oracle}}$}: $\sum_{i=1}^{N}\frac{1}{m}\|u_i-Af_\theta(y_i)\|^2 + \alpha\mathcal{L}_{\text{EQ}}(\theta)$

    \item \textbf{$\text{REI}_{\text{oracle}}$}: $\sum_{i=1}^{N}\frac{1}{m}\|u_i-Af_\theta(y_i)\|^2 +\alpha \mathcal{L}_{\text{REQ}}(\theta)$

    \item \textbf{$\text{REI}$}: $\mathcal{L}_{\text{SURE}}(\theta) +\alpha \mathcal{L}_{\text{REQ}}(\theta)$
    \item \textbf{Sup}:  $\sum_{i=1}^{N}\frac{1}{n}\|x_i-f_\theta(y_i)\|^2$
\end{itemize}

Figure~\ref{fig:ei_variants} shows the reconstruction performance (PSNR) as a function of noise level $\sigma$ for different variants of training loss on the $4\times$ accelerated MRI image reconstruction task. Recall the oracle losses include additional oracle access to the clean measurements $u_i$. This provides us with a way to study the contributions of the equivariant and SURE
losses.

From the figure we can observe that: (i) MC fails to learn anything except to converge to the linear reconstruction $\eAT y$, while SURE achieves a stable estimation to the clean measurement consistency. (ii) REI outperforms $\text{EI}_1$ and $\text{EI}_2$ which demonstrates the effectiveness of the proposed training loss (Equation (14) in the main paper). (iii) REI performs better than $\text{EI}_{\text{oracle}}$ demonstrating the benefits of using a noisy input in our proposed robust Equivariance loss. (iv) Both Sup and $\text{REI}_{\text{oracle}}$ outperform REI due to having access to the ground truth clean images and measurements, respectively. However, as noted in the main paper the REI performance lies close to that for $\text{REI}_{\text{oracle}}$, suggesting that the SURE loss is doing a reasonable job of estimating the (oracle) clean measurement consistency loss.

\subsection{Effect of the equivariance hyperparameter $\alpha$}
We consider the optimal value and sensitivity of the hyperparameter $\alpha$ on the inpainting task.
Table~\ref{table:alpha} shows the REI reconstruction performance for different equivariance strength values ($\alpha$ in Equation (14) of the main paper) and different Poisson noise levels $\gamma$.
We see that an optimal value here is around $\alpha=1$, although the performance is generally good over the range $0.1 \leq \alpha \leq 1$ indicating that REI is not sensitive to the precise value of $\alpha$. However, when $\alpha$ is either too small or too large we do observe deterioration in performance.
%It performs reasonably well on all the range of noises when $\alpha=1$. When $\alpha$ is too small, the performance drops considerably and fails to learn the nullspace well. When $\alpha$ is too big, the performance drops rapidly due to less noise elimination.
Consistent observations were also found in the MRI task. Therefore, we set $\alpha=1$ for inpainting and MRI tasks.

In the nonlinear CT task, since the measurement values are substantially larger than those in the MRI and inpainting experiments we scaled the SURE loss by $10^{-5}$ and set $\alpha=1000$. We leave further optimization of $\alpha$ in this task for future research.

\begin{table}[h]
\begin{center}
\fontsize{7}{12}\selectfont
\begin{tabular}{c|ccccc}
$\gamma$&$\alpha=0.01$&  $\alpha=0.1$ & $\alpha=1$ & $\alpha=10$ &$\alpha=100$\\
\hline
$0.01$ &19.5 $\pm$ 1.1 & 19.66 $\pm$ 1.7 & 21.0 $\pm$ 1.3 & 16.2 $\pm$ 1.6 & 10.6 $\pm$ 1.4\\
$0.05$ &17.0 $\pm$ 1.0 & 18.1 $\pm$ 1.0 & 18.2 $\pm$ 1.2 & 13.5 $\pm$ 1.7 & 7.4 $\pm$ 2.0\\
$0.1$ &11.2 $\pm$ 1.1 & 16.0 $\pm$ 1.0 &  16.6 $\pm$ 1.3 & 12.3 $\pm$ 1.5 & 8.0 $\pm$ 1.6
\end{tabular}
\end{center}
% \vspace{-10pt} # only for non-anonymous
\caption{Effect of the equivariance strength $\alpha$ on the reconstruction performance (PSNR) in the inpainting reconstruction (Urban 100 dataset) task with different noise level $\gamma$.}\label{table:alpha}
\end{table}

\subsection{Effect of the small positive number $\tau$}
%In general, the value of $\tau$ is critical to the SURE-based method.
We follow the suggestion of \cite{metzler2018unsupervised} and select the value of $\tau$ to be around $\text{max}(y)/1000$. Table~\ref{table:tau} presents REI reconstruction performance (PSNR) on the MRI task with different magnitudes of $\tau$ where the Gaussian noise level is fixed as $\sigma=0.1$. It shows that REI works best when $\tau=10^{-2}$. In the MRI and inpainting experiments, we set $\tau=10^{-2}$ and set $\tau=10$ to the CT task to account for the larger values of $y$.

\begin{table}[h]
\begin{center}
\fontsize{7}{12}\selectfont
\begin{tabular}{c|cccc}
$\sigma$&$\tau=10^{-1}$&  $\tau=10^{-2}$ & $\tau=10^{-3}$ & $\tau=10^{-4}$\\
% $\sigma$&$\tau=0.1$&  $\tau=0.01$ & $\tau=0.001$ & $\tau=0.0001$\\
\hline
$0.1$ &24.6 $\pm$ 2.2 & 27.7 $\pm$ 2.0 & 26.9 $\pm$ 2.3 & 25.3 $\pm$ 2.4
\end{tabular}
\end{center}
% \vspace{-10pt} # only for non-anonymous
\caption{Sensitivity of the small positive number $\tau$ on the reconstruction performance (PSNR) in the $4\times$ accelerated MRI (fastMRI dataset) task with a fixed noise level $\sigma=0.1$.}\label{table:tau}
\end{table}

\begin{table}[h]
\begin{center}
\fontsize{7}{12}\selectfont
\begin{tabular}{c|ccccc}
&FBP&  EI & REI & $\text{REI}_{\text{oracle}}$ &Sup\\
\hline
CT &12.9 $\pm$ 1.1 & 21.6 $\pm$ 0.7 & 30.5 $\pm$ 1.0 & 30.5 $\pm$ 1.1 & 30.7 $\pm$ 1.0
\end{tabular}
\end{center}
% \vspace{-10pt} # only for non-anonymous
\caption{Reconstruction performance (PSNR) of \emph{low-dose} 50-views CT (CT100 dataset) task for different methods on the test noisy measurements. $I_0=10^4$ and $\sigma=50$.}\label{table:ct_sm}
\end{table}

\subsection{Low-dose Sparse-view MPG CT}
We have shown in the main paper an initial demonstration that REI can learn a competitive CT reconstruction with MPG noisy measurements in the nonlinear CT task. Here we are interested in further testing whether REI can handle the more challenging low-dose (i.e., higher Poisson noise) case. To do that, we  set $I_0=10^{4}$ and $\sigma=50$, Table~\ref{table:ct_sm} and Figure~\ref{fig:results_ct_sm} show some preliminary results. We have the following observations: (i) both FBP and EI fail to learn the reconstruction due to the high MPG noise in the measurements.  (ii) In some examples (e.g., the first row in  Figure~\ref{fig:results_ct_sm}), REI enjoy a slightly higher PSNR than that of Sup, but visually the reconstruction result of Sup is sharper. This may indicate that the Sup network is overfitting and that REI enjoys better generalization on the test measurements due to the equivariance constraint, c.f EI in~\cite{chen2021equivariant}. (iii) As shown in Table~\ref{table:ct}, REI achieved comparable performance (PSNR) to both Sup and $\text{REI}_{\text{oracle}}$. Moreover, our model outperforms FBP by more than 17 dB and enjoys a 9 dB gain against EI. Both these results and those in our main paper here suggest that robust equivariant imaging is a powerful learning paradigm. We plan to explore the nonlinear CT further in future research, e.g., a wider range of noise levels, hyperparameter optimisation, etc.

\end{document}